\documentclass[10pt,twocolumn,letterpaper]{article}

\usepackage{cvpr}              
\definecolor{cvprblue}{rgb}{0.21,0.49,0.74}
\usepackage[pagebackref,breaklinks,colorlinks,allcolors=cvprblue]{hyperref}
\usepackage{multirow} 
\usepackage{adjustbox} 
\usepackage{siunitx} 
\usepackage{xcolor} 
\usepackage[table]{xcolor}
\usepackage{subcaption}

\def\paperID{1707} 
\def\confName{CVPR}
\def\confYear{2026}

\title{LASAR: Towards Spatio-temporal Reasoning with Latent Cognitive Map}

\author{
  Jinzhou Tang$^{1}$\thanks{Equal Contribution},
  Sidi Liu$^{1*}$,
  Waikit Xiu$^{1*}$,
  Weixing Chen$^{1}$,
  Keze Wang$^{1}$\thanks{Corresponding Author: \tt{kezewang@gmail.com} \\ \quad Code: \url{https://github.com/tangjzh/LASAR}}
  \\
  \\
  $^{1}$Sun Yat-sen University
}

\makeatletter
\patchcmd{\@maketitle}
  {\end{center}}
  {%
    \end{center}%
    \vspace{-4em}%
    \begin{center}%
      \includegraphics[width=\linewidth]{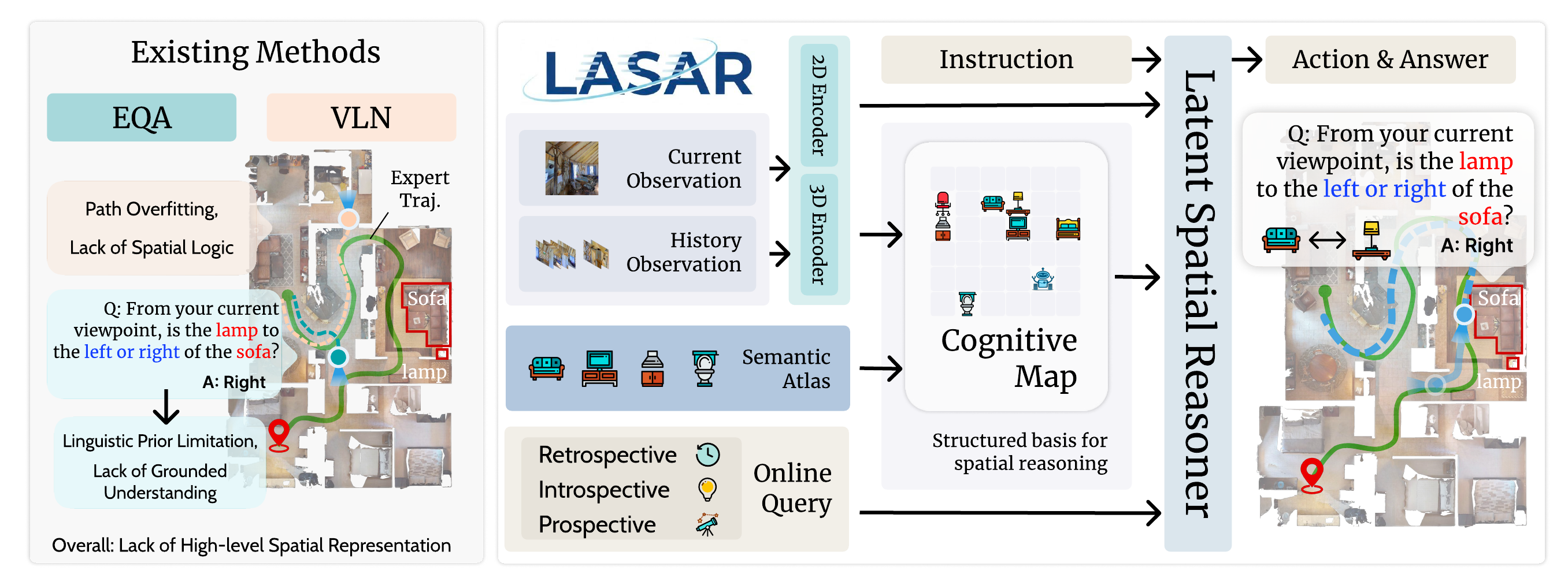}\par%
      \vspace{-1em}%
      \captionof{figure}{We introduce LASAR, a framework that builds a dynamic Cognitive Map from instructions and observations to capture spatial semantics and relations. This map provides structured a basis for spatial reasoning, allowing a Latent Spatial Reasoner to \textit{reason in space}.}%
      \label{fig:intro}
      \vspace{-0.5em}%
    \end{center}%
  }
  {}{}
\makeatother

\begin{document}
\maketitle
\begin{abstract}
A fundamental challenge in embodied AI is verifying if agents build internal models of spatial structure or merely learn to mimic task-specific expert trajectories. This is critical as foundational approaches rooted in action-centric tasks (e.g., VLN) and reasoning-centric tasks (e.g., EQA) often share a common limitation: they lack a learning signal that forces them to encode fine-grained spatial relationships (like topology or distance) over long-range, fragmented experiences. To address this, we first propose LASAR, an architecture featuring a dual-memory system designed to maintain both episodic experiences and a semantic cognitive map. We then introduce Spatio-temporal Contextual Representation Learning (ST-CRL), a contrastive objective designed to train this architecture. ST-CRL leverages spatio-temporal cues from cognitive queries generated through annotated spatio-temporal context in simulation to build sample pairs, thereby forming the internal cognitive map from the agent's experiences. Experiments demonstrate that our method achieves 2\%-3.5\% gains in both zero-shot generalization on standard VLN-CE and VSI-Bench benchmarks. We also demonstrate that our proposed cognitive map has high self-consistency. 
\end{abstract}     
\section{Introduction}
\label{sec:intro}


A fundamental challenge in Embodied AI~\cite{octo_2023} is to build agents that move beyond mimicking superficial policies and instead develop a robust, generalizable understanding of their 3D world~\cite{zou20253d,10.3389/fncom.2020.00063,COPPOLINO202397,Epstein2017TheCM,10.1609/aaai.v38i7.28597}. Current research has progressed along a spectrum spanning action-centric tasks and reasoning-centric tasks. At one end, Vision-Language Navigation (VLN)~\cite{anderson2018vision} challenges an agent to execute a sequence of actions to reach a goal based on instructions, thus prioritizing decision-making and action. At the other, Embodied Question Answering (EQA)~\cite{DBLP:journals/corr/abs-1711-11543} requires an agent to answer complex questions about an environment, thereby focusing on understanding and reasoning~\cite{DBLP:journals/corr/abs-1912-11121,zhu2023ponderv2,huang2023ponder,yang2023unipad}.

However, the traditional distinction between ``action" and ``reasoning" highlights critical challenges. On the action-centric side, it remains unclear whether agents trained on massive vision-language-action pairs~\cite{10.5555/3618408.3618748} truly understand space or merely overfit to superficial statistical biases within expert trajectories~\cite{DBLP:journals/corr/abs-1905-12255}. Many studies confirm that imitation learning alone is insufficient for this deep understanding~\cite{DBLP:journals/corr/abs-1901-03035}. Similarly, on the reasoning-centric side, techniques (\textit{e.g.}, CoT~\cite{wei2022chain}) that rely on large language models' linguistic priors also fail at complex spatial tasks~\cite{yang2025thinking}, as their abstract reasoning is detached from a grounded world model, which formal work demonstrates is essential for robust understanding and prediction~\cite{DBLP:journals/corr/abs-1803-10122}. These discoveries highlight the need for a spatial-oriented world model to comprehend the environment. While recent work has largely focused on bridging this gap by combining VLN and EQA tasks~\cite{zhou2024navgpt,zhou2024navgpt2,DBLP:conf/rss/ZhangWXZHF0ZW24}, it remains largely unexplored how this combination can be leveraged to form temporally consistent, high-level spatial representations (\textit{e.g.}, topology, distance, and relations) that enable robust generalization across novel environments and instructions. To this end, we argue that learning a cognitive map~\cite{DBLP:journals/corr/abs-2005-12256,xiang2023language}, which acts as the module that converts raw experience (i.e., a fragmented stream of \{\textit{observation, action}\} pairs) into a queriable world model~\cite{berg2025semantic,wang2025remi}, is a promising approach to building the more robust, high-level spatial logic required for spatio-temporal reasoning.

To address this, we introduce \textbf{LASAR} (\textbf{LA}ten \textbf{S}p\textbf{A}tial \textbf{R}easoner), an LLM-based agent designed to maintain two distinct forms of memory: (1) an episodic memory that stores the sampled temporal sequence of visual-spatial features from the agent's past observations, and (2) a semantic memory that forms a structured latent cognitive map by querying a learnable codebook of world primitives (termed `Spatial Semantic Atlas') using the agent's episodic memory. These two memory systems interact dynamically: the episodic memory provides temporal context, while the semantic memory provides spatial relations. A unified LLM backbone reasons over both memories to integrate task queries, ground task-oriented objects, and generate linguistic responses for downstream tasks. To ensure this latent map encodes robust spatial logic, we propose our core innovation, Spatio-temporal Contextual Representation Learning (ST-CRL), to train it. ST-CRL is a contrastive objective that shapes the semantic memory by requiring the agent to answer challenging cognitive queries. These queries are enabled by our proposed MindCraft task, which introduces Q\&A concurrently with navigation. We generate the corresponding data via a procedural pipeline built on VLN-CE that injects retrospective, introspective, and prospective queries. ST-CRL then leverages these queries and the simulator's privileged information to build high-level spatial-semantic contrastive sample pairs, enabling the training of a generalizable cognitive map for downstream tasks.




Experimental results show a 3.5\% absolute improvement on the VSI-Bench zero-shot reasoning benchmark and improve SPL, SR on VLN-CE by 2\%-5\% over SOTA baselines. Our main contributions are threefold: (1) We propose LASAR, a novel LLM-based agent \textit{reasons in space} through a latent cognitive map. (2) We propose ST-CRL, an objective to sculpt this cognitive map, distilling spatio-temporal logical structure from fragmented embodied experiences. (3) We introduce MindCraft, a task framework and related dataset that unifies action and reasoning by injecting a concurrent stream of cognitive queries into navigation.


\section{Related Work}
\label{sec:related}

\paragraph{Embodied Navigation and Reasoning}

Early embodied agents leveraged reinforcement learning (RL)~\cite{wang2024research,chen2020mapbased,pmlr-v229-liang23a} to acquire navigation skills, with policy optimization driven by interactions with the environment. The emergence of Visual-Language Navigation (VLN)~\cite{liu2024volumetric,zheng2024towards} represents a significant advancement, enabling agents to follow natural language instructions, thereby providing greater flexibility and generalization in task descriptions. With the development of large multimodal models, the field of Embodied Question Answering (EQA)~\cite{majumdar2024openeqa,yu2025seqafford,mu2023embodiedgpt} has emerged in parallel, aiming to assess an agent's understanding of its environment, particularly focusing on spatial perception and reasoning abilities. However, for a long time, the technical solutions for both areas have relied on specialized model architectures and optimization pathways, failing to form effective synergies. To address this issue, integrating the long-term action planning of VLN with the deep reasoning capabilities of EQA can create a framework that simultaneously supports instruction navigation and environmental understanding reasoning, ultimately breaking down technical barriers and achieving a deep fusion of both capabilities.

\paragraph{Memory and Internal Representations}
To succeed in long-horizon tasks, an agent requires robust memory. One line of work focuses on explicit memory, using techniques like Simultaneous Localization and Mapping (SLAM) to build precise 3D geometric maps of the environment~\cite{jia2022learning, ji2024neds, sun2024high}. An alternative is implicit memory~\cite{wang2025causality,jain2023mnemosyne}, where recurrent or transformer-based models encode an agent's history into a compact state vector~\cite{hong2021vlnbert}. More recently, research\cite{wu2025spatial} has begun to explore the implicit cognitive maps that emerge within large multimodal models. A persistent challenge across these paradigms is the difficulty of forming a globally consistent representation from a stream of egocentric views. Indeed, recent benchmarks have quantified this issue, revealing that even powerful models excel at local spatial awareness but struggle with long-range spatial relationships, which suggests a need for novel methods to better guide the construction of these internal maps.

\begin{figure*}[!ht]
    \centering
    \includegraphics[width=\textwidth]{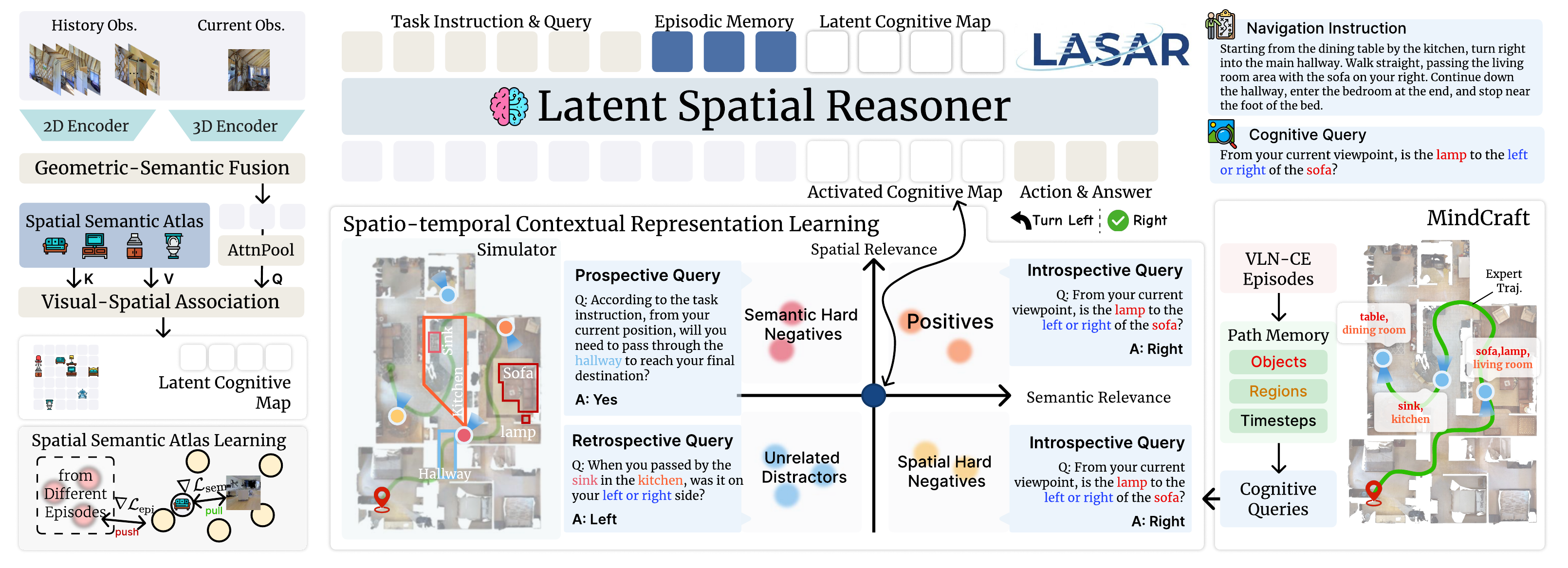}
    \vspace{-25pt}
    \caption{\textbf{Overview of LASAR.} Observations are processed by 2D and 3D Encoders and merged via Geometric-Semantic Fusion for better spatial understanding. These features populate a dual-memory system: (1) a high-fidelity Episodic Memory storing the temporal visual history, and (2) a structured Latent Cognitive Map. This map is generated by querying a learnable codebook (\textit{i.e.}, the Spatial Semantic Atlas) with episodic context via cross-attention (\textit{i.e.}, Visual Spatial Association). The Latent Spatial Reasoner (an LLM backbone) integrates the Task Instruction \& Query with both memory streams; its key function is to leverage the LLM's powerful contextual understanding to dynamically refine the general Latent Cognitive Map ($m_t$) into a task-centric, Activated Cognitive Map ($m_t^{\prime}$) required to answer the specific query. The framework is trained using the MindCraft pipeline (right), which injects Cognitive Queries into VLN-CE Episodes for valid timesteps. These queries provide the supervisory signal for our core Spatio-temporal Contextual Representation Learning (ST-CRL) objective (center), which shapes the latent space by constructing contrastive sample pairs for the cognitive map.}
    \vspace{-15pt}
    \label{fig:overview}
\end{figure*}

\paragraph{Spatio-temporal Representation Learning}
Spatio-temporal representation learning is fundamental for building predictive world models~\cite{huang2021spatio,zhao2023generative,zhang2024multi}. In egocentric video, self-supervised learning (SSL) has been effective, often using predictive tasks~\cite{qi2021self} or contrastive objectives based on temporal proximity or data augmentations~\cite{planamente2021self,akiva2023self}. While these methods learn powerful representations for action-centric tasks by capturing short-term dynamics, their supervisory signals are derived from raw sensory data. Consequently, they are often agnostic to the agent's high-level cognitive state, and the resulting representations are not explicitly optimized for complex, long-range spatial reasoning. Our work, ST-CRL, addresses this gap by using spatio-temporal cognitive queries as a novel, high-level supervisory signal to directly sculpt a reasoning-aware representation.

\section{Methodology}
\label{sec:method}


\subsection{Problem Formulation}

\paragraph{MindCraft Task Definition.}

An agent interacts with a 3D environment $\mathcal{E}$. At each timestep $t$, the agent at pose $p_t$ receives a visual observation $o_t$ and must select an action $a_t \in \mathcal{A}$. Our proposed task, \textbf{MindCraft}, introduces a concurrent online querying mechanism atop this standard process. Given an instruction $\mathcal{T}$, the agent must learn a unified policy $\pi$. At any timestep $t$, the policy receives the history $H_t$ and an optional query $q_t \in \mathcal{Q} \cup \{\emptyset\}$. The policy must always output a navigation action $a_t$ and, if a query $q_t$ is present, additionally output an answer $ans_t$:
\begin{equation}
\pi(H_t, \mathcal{T}, q_t) \rightarrow (a_t, ans_t)
\end{equation}
To systematically probe the agent's cognitive map, queries $q_t$ are categorized into a three-tiered hierarchy: (1) \textbf{Retrospective Queries}, which probe the agent's memory of past observations (\textit{i.e.}, Object Attribute Recall and Temporal Relation Recall); (2) \textbf{Introspective Queries}, which probe the agent's understanding of its current state (\textit{i.e.}, Self-Localization and Local Spatial Relation); and (3) \textbf{Prospective Queries}, which probe the agent's ability to predict or plan (\textit{i.e.}, Topological Adjacency Prediction and Future Landmark Prediction)~\cite{DBLP:journals/corr/abs-1711-11543}.

\paragraph{MindCraft Dataset Generation}

Our procedural pipeline for generating the MindCraft dataset is built upon the existing \textbf{Vision-and-Language Navigation in Continuous Environments (VLN-CE)} dataset, as illustrated on the right of Figure~\ref{fig:overview}. We leverage its provided expert trajectories and high-fidelity 3D environments (\textit{e.g.}, Matterport3D) to inject a concurrent, online cognitive reasoning dimension into the classic navigation task. Further implementation details for each query generation module are detailed in Supp.~\ref{app:dataset}.

\subsection{LASAR: Latent Spatial Reasoner}

As shown in Figure~\ref{fig:overview}, LASAR is built on a dual-memory system: an episodic memory that records past observations, and a visual-spatial association module that connects these observations with a spatial semantic atlas to produce a structured, latent cognitive map for LLM Reasoning.

\paragraph{Input Encoders.}
The model's front-end consists of a series of frozen pretrained encoders for multi-modal inputs. For each visual observation $o_t$, we extract two complementary feature streams. (1) Standard visual semantic features, $F_{\text{vis},t}$, are extracted from the RGB image using a conventional vision encoder (\textit{i.e.}, Siglip~\cite{tschannen2025siglip}). (2) Concurrently, rich geometric features, $F_{\text{geo},t}$, are inferred from the same RGB image using a module that is pretrained on a large-scale point cloud prediction task, endowing it with the ability to extract 3D features directly from 2D visual cues (\textit{i.e.}, VGGT~\cite{wang2025vggt}). These features provide crucial, view-dependent 3D structural priors \textit{(e.g.}, depth and layout representations) that are complementary to the 2D semantic features from $F_{vis,t}$. These two feature streams are then fused via a learnable 8-head cross-attention~\cite{c:22} layer $\text{CrossAttn}(Q, K, V)$ with a residual connection to produce a semantic-rich, geometry-aware visual representation $F'_{\text{vis},t}$:
\begin{equation}
    F'_{\text{vis},t} = F_{\text{vis},t} + \text{CrossAttn}(F_{\text{vis},t}, F_{\text{geo},t},F_{\text{geo},t})
\end{equation}

\paragraph{Episodic Memory.}
We implement episodic memory, $M_{\text{epi}, t}$, as a direct record of the agent's experiences, \textit{i.e.}, a temporally ordered sequence of all past geometry-aware visual representations:
\begin{equation}
    M_{\text{epi}, t} = (F'_{\text{vis},0}, F'_{\text{vis},1}, ..., F'_{\text{vis},t})
\end{equation}

\paragraph{Spatial Semantic Memory.}
This memory system generates the agent's cognitive map. It consists of two key components. The foundation is a spatial semantic atlas, $E_{\text{world}} = \{e_1, ..., e_{N_w}\}$, which is a learnable codebook~\cite{DBLP:journals/corr/abs-1711-00937} of general world primitives with semantics and spatial cues (\textit{e.g.}, lamp near sofa, sink in the kitchen). The map itself, which we call the Latent Cognitive Map ($m_t$), is a single vector representation generated in real-time. This map $m_t$ is formed by querying the atlas using the agent's current context.

To achieve this, the data flows as follows: first, an attention pooling layer summarizes the entire episodic memory $M_{epi,t}$ into a single context vector $z_t$. This vector $z_t$ then acts as the query to the Atlas ($E_{\text{world}}$), which provides the keys and values for a cross-attention operation:
\begin{equation}
    m_{t} = \text{CrossAttn}(z_t, E_{\text{world}}, E_{\text{world}})
\end{equation}

\paragraph{Unified Decision Head.}
All components of the LASAR architecture converge into a unified decision-making head, which comprises an LLM backbone responsible for integrating all information and generating the final output. Specifically, at timestep $t$, the LLM backbone input $\mathcal{T}_{\text{in}}$ to generate $(a_t,ans_t)$:
\begin{equation}
    \mathcal{T}_{\text{in}} = [H_{\mathcal{T}} ; M_{\text{epi}, t} ; m_{t}]
\end{equation}
where $H_{\mathcal{T}}$ represents the feature embedding of the language instruction $\mathcal{T}$. To manage computational complexity with long trajectories, we employ a frame sampling strategy following LongVILA, where we uniformly sample a fixed number of frames (\textit{i.e.}, 32) from the entire history $M_{epi,t}$.

We input both $M_{\text{epi}, t}$ and $m_t$ as they provide complementary representations at different levels of abstraction, which are critical for a robust hierarchical reasoning process: (1) The Latent Cognitive Map ($m_t$) serves as a structured, abstract spatio-temporal representation, encoding relational knowledge that acts as a cognitive index. Its lower dimensionality provides the LLM with an efficient, low-cost overview of global context (\textit{i.e.}, ``where am I?" and ``what is around me?"). (2) The Episodic Memory ($M_{\text{epi}, t}$) provides the high-fidelity, unsummarized temporal record of past observations, serving as the grounding evidence.

This dual-input strategy allows the LLM's attention mechanism to first leverage the latent cognitive map $m_t$ to efficiently contextualize a query by locating the relevant spatio-temporal region, and second, to attend precisely to the specific, fine-grained details within the spatio-temporal context in $M_{\text{epi}, t}$ to verify facts and recall precise information. To handle the dual outputs of navigation action $a_t$ and answer $ans_t$, we adopt a unified vocabulary approach. The navigation action space $\mathcal{A}$ (\textit{e.g.}, move forward 75cm, turn left 15 degrees) is treated as special tokens~\cite{brohan2023rt2visionlanguageactionmodelstransfer} within the VLM's vocabulary. Depending on whether a query $q_t$ is present, the model is prompted to autoregressively generate either an action token or a sequence of text for the answer.

\subsection{Training Objectives}

\paragraph{Main Task Loss.} The main task loss at timestep $t$ is composed of two parts: the action prediction loss ($\mathcal{L}_{\text{action}, t}$) via imitation learning, and the query answering loss ($\mathcal{L}_{\text{QA}, t}$) for sequence generation.
\begin{equation}
\mathcal{L}_{\text{MindCraft}, t} = \mathcal{L}_{\text{action}, t} + \lambda_{qa} \cdot I(q_t \neq \emptyset) \cdot \mathcal{L}_{\text{QA}, t}
\end{equation}

To ensure the dual-memory system develops a truly structured cognitive map, we introduce three auxiliary losses.

\paragraph{Spatio-temporal Contextual Representation Learning.}
The cornerstone of our training strategy is the Spatio-temporal Contextual Representation Learning (ST-CRL) loss. This contrastive objective is designed to ensure the Latent Cognitive Map ($m_t$) encodes fine-grained relational information, moving beyond simple co-occurrence statistics.

To achieve this, the loss is not applied to $m_t$ directly, but to a query-conditioned representation, $m'_t$, which is generated by the LLM backbone itself. The logic is that by forcing the LLM's query-conditioned output ($m'_t$) to be spatially and semantically consistent, the gradients from this loss will flow back through the LLM and update the components that produce the original $m_t$ (\textit{i.e.}, the spatial semantic atlas $E_{\text{world}}$). This process compels $m_t$ to be structured in a way that is optimal for downstream reasoning.

We generate this query-conditioned anchor $m'_t$ as follows: First, the Latent Cognitive Map $m_t$ is injected into the LLM's input sequence. We achieve this by adding a new special token, \texttt{[MAP]}, to the LLM's vocabulary and deterministically replacing its embedding with the $m_t$ vector at each forward pass. The LLM's forward pass yields the Activated Cognitive Map, $m'_t$, extracted from the final hidden layer at the \texttt{[MAP]} token's position. This vector, which represents the map as viewed ``through the lens" of the query, serves as the anchor for our contrastive objective. For a given anchor $m'_t$, we construct a set of positive and hard-negative samples by mining from a replay buffer indexed by the spatio-temporal and semantic metadata programmatically generated with each task instance. 

All sample representations are generated homogeneously by passing their respective experiences through the identical LLM backbone. The positive sample ($m'_p$) is generated from a distinct experience in which the query is semantically equivalent to the anchor's, as determined by a shared identifier from our procedural query generation template, and yields the same ground-truth answer. The negative samples ($\mathcal{N}_t$) are a set $\{m'_n\}$ composed of strategically selected hard negatives: (1) Spatial Hard Negatives, identified as experiences with a semantically equivalent query to the anchor's but yielding a different ground-truth answer, which programmatically finds instances referring to a different state in a disparate spatial context; (2) Semantic Hard Negatives, sampled from the same spatial region as the anchor (sharing the same \texttt{region\_id} provided by the simulator environment) but corresponding to a different query and answer; and (3) Unrelated Distractors, where both the \texttt{region\_id} and the query template identifier differ from the anchor's. The ST-CRL loss is then formulated using the standard InfoNCE objective:
\begin{equation}
\label{eq:st_loss}
\mathcal{L}_{\text{crl}} = \text{InfoNCE}(m'_t, m'_p, \mathcal{N}_t)
\end{equation}
We note that the \texttt{region\_id} provided by the simulator is used only during the training phase as privileged information. This signal is essential for sculpting the latent space by providing high-level spatial context (\textit{e.g.}, ``kitchen" vs. ``bedroom") for the negative sampling process. This information is not available to the model during inference. Key hyperparameters for this objective, such as the number of negative samples and the temperature $\tau$, are detailed in the Supp.~\ref{app:hyperparams}.

\paragraph{Spatial Semantic Atlas Learning.}
The foundation of the cognitive map is the spatial semantic atlas, $E_{\text{world}}$. To ensure its primitives are both representative and diverse, this loss uses vector quantization to encourage the nearest primitive $e_j$ to match the visual feature $F'_{vis,t}$, while an entropy regularizer prevents model collapse.
\begin{equation}
\mathcal{L}_{\text{sem}} = || \text{sg}(F'_{vis,t}) - e_j ||_2^2 - \gamma \sum_{k=1}^{N_w} p_k \log(p_k)
\end{equation}
Here, $p_k$ denotes the probability of the $k$-th atlas primitive being selected as the closest match for any given observation within a training batch. The entropy term thus encourages the distribution of primitive usage to be uniform, preventing the model from relying on a small subset of the atlas. While $\mathcal{L}_{crl}$ structures the atlas for high-level reasoning, $\mathcal{L}_{sem}$ serves as a vital grounding mechanism. It ensures that the atlas primitives $e_j$ remain tethered to and representative of the underlying visual-geometric features $F_{vis,t}^{\prime}$ from which they are derived.



\paragraph{Episodic Discriminability.}
The quality of the cognitive map depends on the quality of the episodic memory it distills from. To ensure the episodic memory $M_{epi}$ effectively encodes trajectory-specific information, we introduce a contrastive loss~\cite{DBLP:journals/corr/abs-1807-03748} at the feature level. The objective is to pull representations from the same episode closer together, while pushing representations from different episodes apart.

This objective forces the encoders to produce features that are discriminative of the specific journey. The loss is formulated as: 
\begin{equation} \mathcal{L}_{\text{epi}} = \text{InfoNCE}(F'^{(i)}_{\text{vis}, t_a}, F'^{(i)}_{\text{vis}, t_p}, {\mathcal{N}}) \label{eq:retro} 
\end{equation}





\paragraph{Final Objective and Inference.}
The final training objective combines all losses. The per-timestep losses are averaged over the trajectory length $T$, and the single per-episode loss ($\mathcal{L}_{\text{epi}}$) is added to this average:
\begin{equation}
\begin{aligned}
\mathcal{L}_{\text{total}} = \frac{1}{T}&\sum_{t=1}^{T} (\mathcal{L}_{\text{MindCraft}, t} + \lambda_{c} \cdot I(q_t \neq \emptyset) \cdot \mathcal{L}_{\text{crl}, t} \\
&+ \lambda_{s}\mathcal{L}_{\text{sem}, t}) + \lambda_{r}\mathcal{L}_{\text{epi}} \label{eq:total_loss}
\end{aligned}
\end{equation}
\section{Experiment}
\label{sec:exp}

\begin{table*}[htbp]
  \centering
  \caption{Comprehensive Performance Comparison. MindCraft evaluates in-domain reasoning under dual load. R2R/RxR evaluates multi-task navigation performance. VSI-Bench evaluates zero-shot reasoning generalization. Note: No models were fine-tuned on VSI-Bench.}
  \vspace{-5pt}
  \label{tab:combined_results}
  \adjustbox{max width=\textwidth}{%
  \begin{tabular}{l | cccc | ccc | cc | ccc}
    \toprule
    \multirow{2}{*}{Model} & \multicolumn{4}{c|}{MindCraft-Test } & \multicolumn{3}{c|}{R2R val-unseen } & \multicolumn{2}{c|}{RxR val-unseen } & \multicolumn{3}{c}{VSI-Bench (Zero-Shot) } \\
    \cmidrule(lr){2-5} \cmidrule(lr){6-8} \cmidrule(lr){9-10} \cmidrule(lr){11-13}
    & QA-Acc $\uparrow$ & GCA $\uparrow$ & CMC $\uparrow$ & SR@WA $\downarrow$ & SR $\uparrow$ & SPL $\uparrow$ & OS $\uparrow$ & SR $\uparrow$ & SPL $\uparrow$ & ACC(MCA) $\uparrow$ & MRA(NA) $\uparrow$ & Avg. $\uparrow$ \\
    \midrule
    \rowcolor[gray]{0.9}
    \multicolumn{13}{l}{\textit{trained w/o. MindCraft}} \\
    GPT-4o~\cite{hurst2024gpt} & 55.2 & - & 22.4 & - & - & - & - & - & - & 33.5 & 36.1 & 34.0 \\
    Gemini-1.5 Pro~\cite{team2024gemini} & 53.5 & - & 20.5 & - & - & - & - & - & - & \textbf{49.7} & 44.0 & 45.4 \\
    NavCoT~\cite{lin2025navcot}  & 40.3 & 48.1 & 60.5 & 59.1 & 40.2 & 36.6 & 48.1 & 24.5 & 22.6 & 34.7 & 29.6 & 32.1 \\
    Uni-NaVid~\cite{zhang2024uni}  & 42.5 & 50.7 & 66.2 & 42.0 & 47.0 & 42.7 & 53.3 & 48.7 & 40.9 & 34.8 & 39.1 & 35.6 \\
    LASAR (IL)  & 56.2 & 57.2 & 65.7 & 56.5 & 54.8 & 52.7 & 61.5 & 50.5 & 44.3 & 40.1 & 35.5 & 37.8 \\
    \midrule
    \rowcolor[gray]{0.9}
    \multicolumn{13}{l}{\textit{trained w. MindCraft}} \\
    NavCoT  & 48.2 & 50.1 & 28.5 & 42.5 & 39.7 & 32.0 & 50.2 & 26.2 & 18.7 & 36.9 & 31.5 & 34.3 \\
    Navid~\cite{zhang2024navid}  & 56.6 & 49.5 & 64.5 & 42.8 & 41.6 & 37.1 & 50.9 & 26.3 & 22.8 & - & - & - \\
    Uni-NaVid  & 54.1 & 52.2 & 68.1 & 40.1 & 48.1 & 44.1 & 55.7 & 50.1 & 42.3 & 37.1 & 42.5 & 39.7 \\
    NaVILA~\cite{cheng2024navila}  & 54.7 & 55.4 & 66.7 & 56.9 & 54.8 & 49.0 & 62.5 & 49.3 & 44.0 & - & - & - \\
    LASAR (IL+QA)  & 60.6 & 63.2 & 70.1 & 57.3 & 55.3 & 47.4 & 62.3 & 49.5 & 43.3 & 42.5 & 47.6 & 44.8 \\
    \textbf{LASAR (Ours)}  & \textbf{65.3} & \textbf{70.4} & \textbf{75.8} & \textbf{35.2} & \textbf{57.0} & \textbf{53.9} & \textbf{63.1} & \textbf{52.1} & \textbf{47.9} & 43.5 & \textbf{52.9} & \textbf{48.9} \\
    \bottomrule
  \end{tabular}
  } 
  \vspace{-10pt}
\end{table*}

This section presents quantitative and qualitative experiments to validate our core hypothesis: that training our model with the proposed ST-CRL objective on the MindCraft task builds a superior latent cognitive map. We aim to demonstrate that this learned representation not only enhances in-domain reasoning capabilities but also acts as a potent auxiliary signal, improving performance and generalization on standard downstream navigation tasks.

\subsection{Experimental Setup}
\label{sec:setup}

\paragraph{Datasets and Tasks}
We train all models on the MindCraft-Train dataset, which is built upon the VLN-CE training set. We evaluate performance across three distinct settings: (1) Evaluate on our MindCraft-Test dataset, which is derived from the VLN-CE val-unseen split. This task requires the agent to perform navigation while concurrently answering cognitive queries. (2) For downstream navigation tasks, to measure the navigation-only benefit, we evaluate on the standard VLN-CE val-unseen~\cite{krantz_vlnce_2020} and RxR val-unseen splits. This isolates the improvement in standard navigation metrics (\textit{e.g.}, SR and SPL) gained from our MindCraft training. (3) For zero-shot reasoning generalization, to test for broader spatial understanding, we evaluate on VSI-Bench~\cite{yang2024think}. This is a video VQA benchmark that the model has never seen during training. Further details on our training data construction are provided in Supp.~\ref{supp:data}.

\paragraph{Evaluation Metrics}
We use metrics specific to each evaluation setting. For the MindCraft dataset, focusing on its unique dual-load nature, we report: (i) Overall Reasoning Accuracy: Query Accuracy (QA-Acc). (ii) Navigation-Conditioned Reasoning: Goal-Conditioned Accuracy (GCA), calculated as the QA-Acc only on trajectories where navigation was successful (SR=1). (iii) Cognitive Map Consistency (CMC), measuring the consistency of answers to semantically equivalent but differently phrased queries targeting the same spatial fact within a trajectory. (iv) Reasoning Failure Impact: Navigation Success Rate on trajectories where at least one query was answered incorrectly (SR@WA). For Downstream Navigation (R2R \& RxR), we use the standard metrics: Success Rate (SR), Success rate weighted by Path Length (SPL), and Oracle Success (OS). For VSI-Bench Zero-Shot, following its standard, we report Accuracy (ACC) for Multiple-Choice Answer (MCA) questions, Mean Relative Accuracy (MRA) for Numerical Answer (NA) questions, and an overall Average score. The precise mathematical formulations for all evaluation metrics are provided in Supp.~\ref{supp:metric}.

\paragraph{Baselines}
We compare LASAR against two categories of models. First, for state-of-the-art agents, we compare against recent strong embodied navigation models, including NavCoT~\cite{lin2025navcot}, Uni-NaVid~\cite{zhang2024uni}, NaVid~\cite{zhang2024navid}, and NaVILA~\cite{cheng2024navila}. These models are retrained on our MindCraft-Train dataset for a fair comparison. We also report zero-shot reasoning scores from foundation models like Gemini-1.5 Pro~\cite{team2024gemini} and GPT-4o~\cite{hurst2024gpt}. Second, to analyze the contributions of our MindCraft dataset, we define two key baseline variants of our model: (1) LASAR (IL), which is our model trained \textit{only} on the imitation learning (action) loss from the navigation task. This serves as our navigation-only baseline. (2) LASAR (IL+QA), which is our model trained on both the imitation loss and the standard query answering loss ($\mathcal{L}_{\text{MindCraft}}$), but \textit{without} our proposed ST-CRL loss or the other auxiliary losses ($\mathcal{L}_{\text{sem}}$, $\mathcal{L}_{\text{epi}}$). This variant allows us to measure the benefit of naively training on the MindCraft task, and it also serves as the direct baseline to isolate the specific contribution of ST-CRL. Implementation details for all baseline models are described in Supp.~\ref{app:benchmark_details}.

\paragraph{Implementation Details} 
Our LASAR architecture employs frozen pre-trained encoders: Siglip~\cite{tschannen2025siglip} for visual semantics ($F_{\text{vis}}$, $D_{\text{vis}}=768$) and a VGGT model~\cite{wang2025vggt} for geometric features ($F_{\text{geo}}$, $D_{\text{geo}}=768$), fused via 8-head cross-attention ($D'_{\text{vis}}=768$). The dual memory consists of Episodic Memory ($M_{\text{epi}}$, sampling $K=32$ frames following LongVILA~\cite{chenlongvila}) and Spatial Semantic Memory ($M_{\text{sem}}$) with a Semantic Atlas ($E_{\text{world}}$) of $N_w=512$ codes ($D_{\text{sem}}=768$). The LLM Backbone uses Qwen2-7B~\cite{yang2024qwen2}. We train using AdamW ($lr=1 \times 10^{-4}$) with a cosine annealing scheduler (warmup ratio $0.03$) for 2 epochs on 8 A100 GPUs (batch size 32) with seed 42. For losses, $\lambda_{qa}=1.0$; auxiliary weights are $\lambda_{c}=0.1$ (ST-CRL, $\tau=0.07$, $N=32$ negatives: 8 spatial, 8 semantic, 16 unrelated), $\lambda_{s}=0.2$ (Semantic Atlas), and $\lambda_{r}=0.1$ (Episodic). Our end-to-end training strategy is detailed in Supp.~\ref{supp:strategy} and ~\ref{app:architecture_details}. And an analysis of computational costs is presented in Supp.~\ref{supp:cost}.

\begin{figure*}[ht!]
    \centering
    \includegraphics[width=\textwidth]{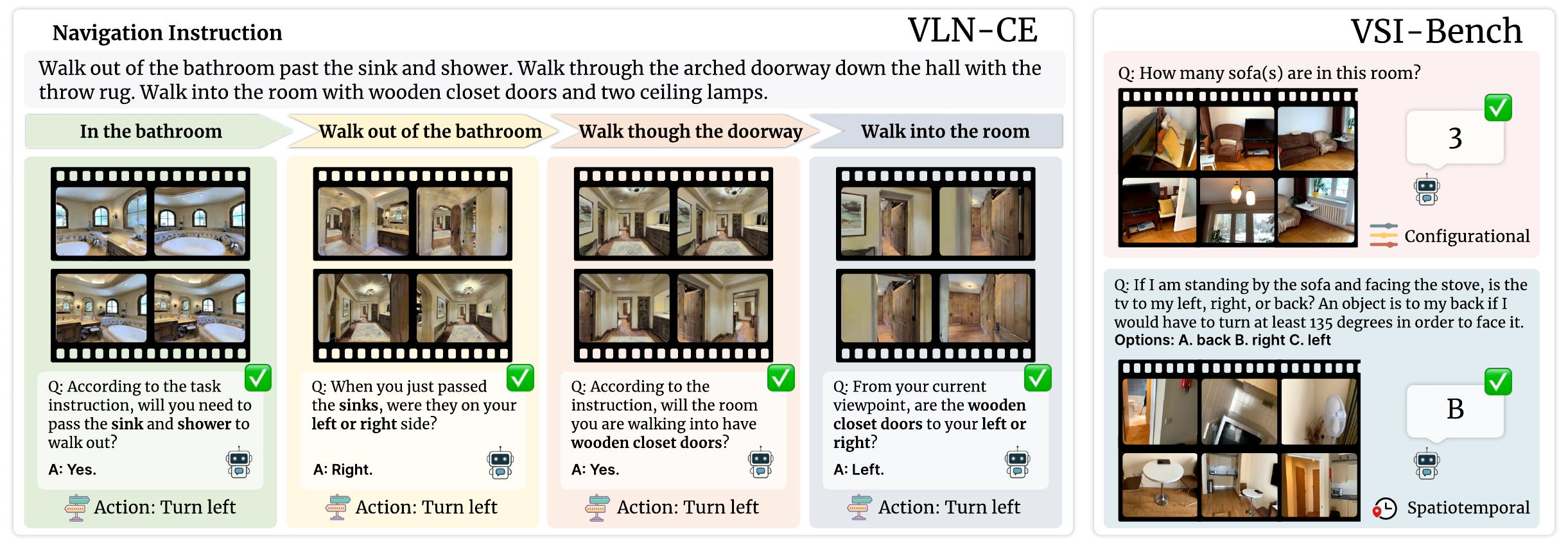}
    \vspace{-15pt}
    \caption{\textbf{Qualitative results from VLN-CE~\cite{krantz_vlnce_2020} and VSI-Bench~\cite{yang2024think}.} We deploy LASAR in the simulation environment (\textit{i.e.}, Habitat) for long-horizon navigation task (\textit{i.e.}, VLN-CE) and Visual-spatial EQA task (\textit{i.e.}, VSI-Bench). We also provide qualitative results of our proposed MindCraft-Test with cognitive queries during navigation.} 
    \label{fig:visualization}
    \vspace{-15pt}
\end{figure*}

\subsection{Main Results and Analysis}
\label{sec:main_results}

Table~\ref{tab:combined_results} presents the performance comparison across MindCraft, VLN-CE evaluation, and VSI-Bench zero-shot generalization for the main models.

\paragraph{MindCraft Performance}
Based on the core MindCraft metrics in Table~\ref{tab:combined_results}, our full model, LASAR (Ours), demonstrates significant performance improvements over the naive multi-task baseline, LASAR (IL + QA). Specifically, our model achieves a QA-Acc of 65.3\%, an absolute gain of 4.7 percentage points over the baseline (60.6\%). Similarly, in goal-conditioned accuracy (GCA), LASAR (Ours) attains 70.4\%, outperforming the baseline by +7.2 percentage points (63.2\%). More importantly, our model exhibits substantially higher cognitive map consistency (CMC) at 75.8\% compared to 70.1\% for the baseline, indicating that the ST-CRL objective facilitates the learning of a more stable internal representation. Additionally, when reasoning fails, the navigation success rate (SR@WA) of our model is considerably lower (35.2\% vs. 57.3\%), suggesting the enhanced cognitive map supports more robust navigation even under reasoning errors.

\paragraph{Navigation and Reasoning Generalization}
We analyze the complementary benefits of the MindCraft task itself versus our ST-CRL method by examining performance on the two downstream task categories, as presented in Table~\ref{tab:combined_results} and Figure~\ref{fig:visualization}.

The R2R and RxR results in Table~\ref{tab:combined_results} clearly show that training with the ST-CRL auxiliary objective enhances performance on the primary navigation task across both splits. While LASAR (IL + QA) performs comparably to the navigation-only NaVILA, our LASAR demonstrates significant improvements. On the R2R dataset, it achieves an increase of +2.2\% in SR, +4.9\% in SPL, and +0.8\% in OS compared to NaVILA. For the RxR dataset, LASAR shows enhancements of +2.8\% in SR and +3.9\% in SPL relative to NaVILA. This supports our hypothesis that the structured spatial representation learned via ST-CRL provides a better foundation for the navigation policy itself, improving not only goal achievement but also path quality, and generalizing across different instruction types and environments (R2R vs. RxR).

The zero-shot evaluation results on VSI-Bench, as shown in Table~\ref{tab:combined_results}, indicate that LASAR(Ours) has strong generalization capabilities, achieving an average score of 48.9, surpassing other LLM-based VLN methods. In contrast, the naive LASAR (IL+QA) baseline achieves 44.8 Avg, while the navigation-only LASAR (IL) performs even lower at 37.8 Avg. This stark difference suggests that LASAR (IL + QA) likely overfitted to the specific query formats in MindCraft, whereas LASAR (Ours), guided by ST-CRL, developed a more fundamental and transferable understanding of spatial concepts like spatial topologies and relations.

\begin{table}[htbp]
  \centering
  \caption{LASAR ablation study on MindCraft-Test. $\Delta$ indicates the change relative to LASAR (Ours). ``w/o. Sem" removes the Spatial Semantic Memory component. ``Aux" refers to the Semantic Atlas ($\mathcal{L}_{\text{sem}}$) and Episodic Discriminability ($\mathcal{L}_{\text{epi}}$) losses. Metrics correspond to those defined in Sec~\ref{sec:setup}.}
  \label{tab:ablation_results}
  \adjustbox{max width=\linewidth}{
  \begin{tabular}{l c c c c}
    \toprule
    Model & {QA-Acc $\uparrow$} & {GCA $\uparrow$} & {CMC $\uparrow$} & {SR@WA $\downarrow$} \\
    \midrule
    LASAR (Ours) & 65.3 & 70.4 & 75.8 & 35.2 \\ 
    \midrule
    LASAR (w/o. Geo)     & $63.8_{\color{red}-1.5}$ & $62.1_{\color{red}-8.3}$ & $66.3_{\color{red}-9.5}$ & $40.4_{\color{red}+5.2}$ \\ 
    LASAR (w/o. Sem)     & $62.1_{\color{red}-3.2}$ & $65.4_{\color{red}-5.0}$ & $58.2_{\color{red}-17.6}$ & $45.7_{\color{red}+10.5}$ \\ 
    LASAR (w/o. Aux)     & $63.5_{\color{red}-1.8}$ & $67.0_{\color{red}-3.4}$ & $72.9_{\color{red}-2.9}$ & $36.8_{\color{red}+1.6}$ \\ 
    \bottomrule
  \end{tabular}
  }
  \vspace{-10pt}
\end{table}

\begin{figure}[htbp]
  \centering
  \includegraphics[width=\linewidth]{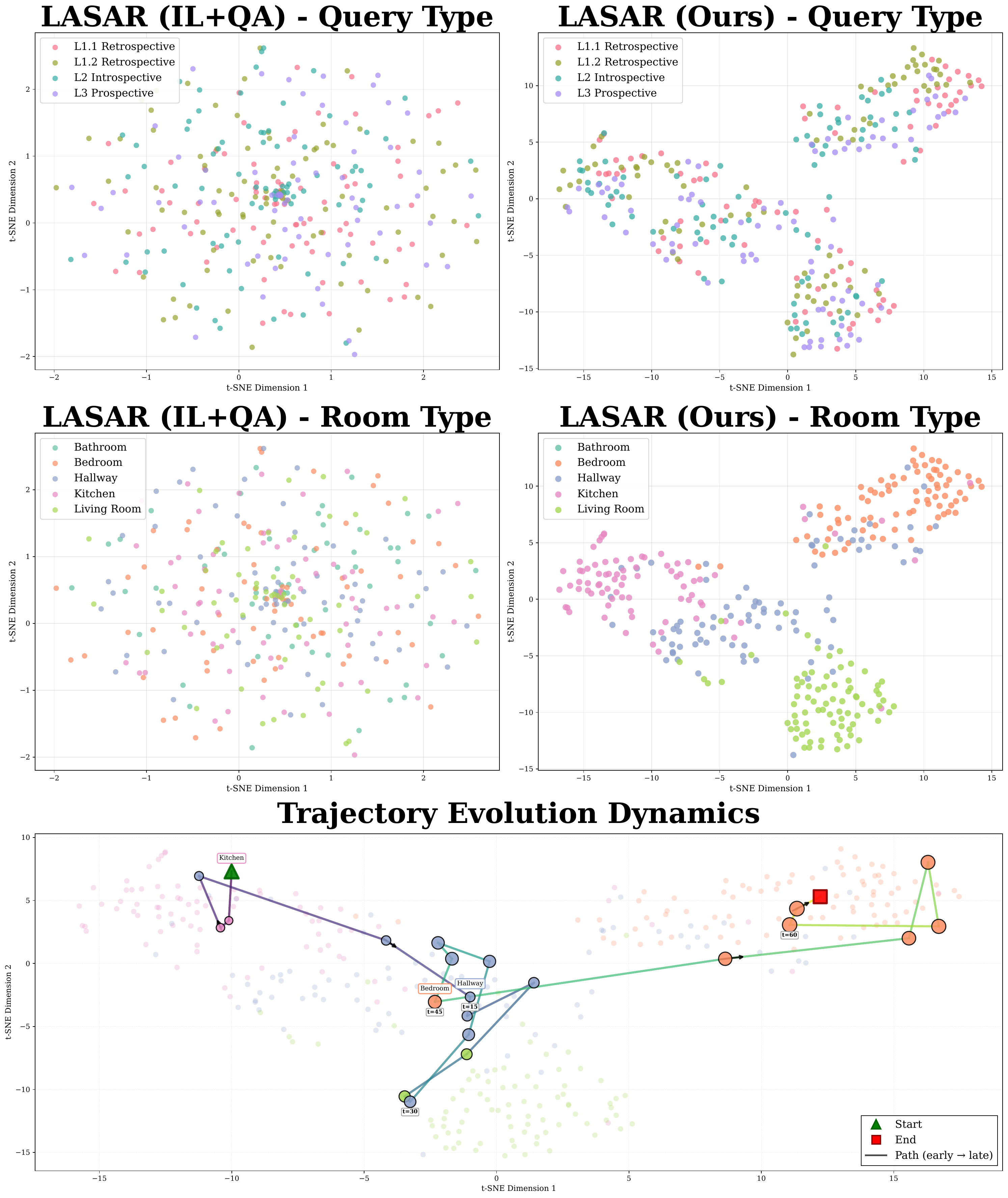} 
  \caption{\textbf{t-SNE visualization of the $m_t$ latent space}, demonstrating the structural impact of ST-CRL.
  We compare the Baseline (LASAR (IL+QA), \textbf{Left Column}) against our full model (LASAR (w. ST-CRL), \textbf{Right Column}).
  \textbf{(Top-Left, Top-Right)} Latent space colored by Query Type.
  \textbf{(Bottom-Left, Bottom-Right)} Latent space colored by Room Type.
  \textbf{(Bottom Row)} Visualization of the $m_t$ state evolution for a single trajectory from our full model.}
\label{fig:tsne}
\vspace{-15pt}
\end{figure}

\subsection{Ablation Study}
\label{sec:ablation}

We conduct an ablation study on the MindCraft-Test dataset to validate our architectural and training objective contributions. The results are presented in Table~\ref{tab:ablation_results}.

\paragraph{Analysis of Training Objectives}
We first isolate the impact of our training objectives. The baseline LASAR (IL+QA) (Table~\ref{tab:combined_results}), which uses only naive imitation and QA loss, achieves a QA-Acc of 60.6\%. Our full model, LASAR (Ours) (Table~\ref{tab:ablation_results}), which adds all three auxiliary losses ($\mathcal{L}_{\text{crl}}$, $\mathcal{L}_{\text{sem}}$, $\mathcal{L}_{\text{epi}}$), achieves 65.3\%. This represents a total gain of +4.7 points. 
We can attribute this gain by analyzing the w/o. Aux model. This model includes ST-CRL ($\mathcal{L}_{\text{crl}}$) but removes the other two auxiliary losses ($\mathcal{L}_{\text{sem}}$, $\mathcal{L}_{\text{epi}}$). It achieves a QA-Acc of 63.5\%, which is only a minor drop (-1.8\%) from the full model. This analysis demonstrates two key findings: (1) The vast majority of the performance gain (4.7 points total) is driven by our core ST-CRL objective, which bridges the gap from 60.6\% (IL+QA) to 63.5\% (w/o. Aux). (2) The remaining auxiliary losses provide a small but beneficial stabilizing effect, adding the final 1.8 points.

\paragraph{Analysis of Architectural Components} We now analyze the core components of our architecture, starting from our full model. (1) Removing the geometric feature fusion (w/o. Geo) leads to a significant drop in reasoning quality, particularly in goal-conditioned accuracy (-8.3\% GCA) and consistency (-9.5\% CMC), despite a smaller drop in overall QA-Acc (-1.5\%). This underscores the value of geometric priors for robust, high-quality spatial reasoning. (2) Removing the entire Spatial Semantic Memory component (w/o. Sem), which replaces the dual-memory structure with only the episodic memory, causes the most severe performance degradation, especially in cognitive consistency (a -17.6\% CMC drop) and a significant fall in overall accuracy (-3.2\% QA-Acc). This confirms that the dual-memory structure, and specifically the latent cognitive map, is critical for handling the complexity of the MindCraft task.

\subsection{Qualitative Analysis}
\label{sec:qualitative}

To qualitatively verify that our ST-CRL objective successfully injects spatial structure into the latent space, we visualize the query-conditioned belief states ($m_t$) using t-SNE (Figure~\ref{fig:tsne}). This visualization primarily demonstrates the macro-level structural impact of our objective.

As shown, the baseline (Left Column) representations are chaotic, unstructured blobs from both the Query Type and Room Type perspectives. This confirms that naive (IL+QA) training fails to organize the latent space. In stark contrast, our full model (Right Column) exhibits clear, semantically meaningful clusters organized by Room Type. This demonstrates ST-CRL's primary success: it leverages spatial context (the \texttt{region\_id}) to sculpt the latent space, providing the necessary high-level topological foundation for reasoning.

While this t-SNE demonstrates the learned macro-level (room) structure, our quantitative metrics provide direct evidence for micro-level (object) relational encoding. The significant gains in CMC (Table~\ref{tab:ablation_results}) and zero-shot VSI-Bench performance (Table~\ref{tab:combined_results}) benchmarks that explicitly test for object-level spatial configuration and relations confirm that this ST-CRL-induced structure successfully embeds the fine-grained spatial relationships required for consistent and generalizable reasoning. Methodological details for this t-SNE visualization are available in Supp.~\ref{supp:featvis}.
\section{Conclusion}

In this work, we argue that the prevalent separation of action-centric (e.g., VLN) and reasoning-centric (e.g., EQA) tasks is a critical barrier to generalizable spatial intelligence. We demonstrate that robust spatial understanding must be learned within a closed loop of action and reasoning. Our proposed MindCraft task framework injects context-aware cognitive probes concurrently with navigation, providing the rich supervisory signal required for our ST-CRL objective. By using the LLM itself as a contextualization engine, ST-CRL learns a structured cognitive map $m_t$ that is not only consistent but also actionable, as demonstrated by its success in guiding navigation policy under constraints. This work provides a concrete methodology for building embodied agents that truly bridge the gap between seeing, reasoning, and acting.

\small
\bibliographystyle{ieeenat_fullname}
\bibliography{main}

@String(CVPR= {IEEE Conf. Comput. Vis. Pattern Recog.})

@String(ECCV= {Eur. Conf. Comput. Vis.})

@String(ICPR = {Int. Conf. Pattern Recog.})

@String(AAAI = {AAAI})

@String(CVPR  = {CVPR})

@String(ECCV  = {ECCV})

@String(ICPR  = {ICPR})

@misc{c:22,
      title={Attention Is All You Need}, 
      author={Ashish Vaswani and Noam Shazeer and Niki Parmar and Jakob Uszkoreit and Llion Jones and Aidan N. Gomez and Lukasz Kaiser and Illia Polosukhin},
      year={2017},
      eprint={1706.03762},
      archivePrefix={arXiv},
      primaryClass={cs.CL}
}

@ARTICLE{10.3389/fncom.2020.00063,
  
AUTHOR={Bermudez-Contreras, Edgar  and Clark, Benjamin J.  and Wilber, Aaron },
         
TITLE={The Neuroscience of Spatial Navigation and the Relationship to Artificial Intelligence},
        
JOURNAL={Frontiers in Computational Neuroscience},
        
VOLUME={Volume 14 - 2020},

YEAR={2020},

URL={https://www.frontiersin.org/journals/computational-neuroscience/articles/10.3389/fncom.2020.00063},

DOI={10.3389/fncom.2020.00063},

ISSN={1662-5188}}

@article{COPPOLINO202397,
title = {An explainable artificial intelligence approach to spatial navigation based on hippocampal circuitry},
journal = {Neural Networks},
volume = {163},
pages = {97-107},
year = {2023},
issn = {0893-6080},
doi = {https://doi.org/10.1016/j.neunet.2023.03.030},
url = {https://www.sciencedirect.com/science/article/pii/S089360802300165X},
author = {Simone Coppolino and Michele Migliore}
}

@article{team2024gemini,
  title={Gemini 1.5: Unlocking multimodal understanding across millions of tokens of context},
  author={Team, Gemini and Georgiev, Petko and Lei, Ving Ian and Burnell, Ryan and Bai, Libin and Gulati, Anmol and Tanzer, Garrett and Vincent, Damien and Pan, Zhufeng and Wang, Shibo and others},
  journal={arXiv preprint arXiv:2403.05530},
  year={2024}
}

@inproceedings{chen2020mapbased,
  title={Robot Navigation with Map-Based Deep Reinforcement Learning},
  author={Chen, Guangda and Pan, Lifan and Chen, Yu'an and Xu, Pei and Wang, Zhiqiang and Wu, Peichen and Ji, Jianmin and Chen, Xiaoping},
  booktitle={Proceedings of the IEEE International Conference on Networking, Sensing and Control (ICNSC)},
  pages={1--6},
  year={2020},
  doi={10.1109/ICNSC48988.2020.9238090}
}

@inproceedings{hong2021vlnbert,
  title     = {VLN‑BERT: A Recurrent Vision‑and‑Language BERT for Navigation},
  author    = {Hong, Yicong and Wu, Qi and Qi, Yuankai and Rodriguez‑Opazo, Cristian and Gould, Stephen},
  booktitle = {Proceedings of the IEEE/CVF Conference on Computer Vision and Pattern Recognition (CVPR)},
  year      = {2021}
}

@misc{
jia2022learning,
title={Learning to Act with Affordance-Aware Multimodal Neural {SLAM}},
author={Zhiwei Jia and Kaixiang Lin and Yizhou Zhao and Qiaozi Gao and Govind Thattai and Gaurav S. Sukhatme},
year={2022},
url={https://openreview.net/forum?id=PtuQ8bk9xF5}
}

@inproceedings{anderson2018vision,
  title     = {Vision-and-Language Navigation: Interpreting visually-grounded navigation instructions in real environments},
  author    = {Anderson, Peter and Wu, Qi and Teney, Damien and Bruce, Joel and Johnson, Mark and Gould, Stephen and van den Hengel, Anton},
  booktitle = {Proceedings of the IEEE/CVF Conference on Computer Vision and Pattern Recognition (CVPR 2018), Spotlight Oral},
  year      = {2018},
  url       = {https://aclanthology.org/2018.CVPR-Spotlight.Anderson/}
}

@InProceedings{pmlr-v229-liang23a,
  title = 	 {Context-Aware Deep Reinforcement Learning for Autonomous Robotic Navigation in Unknown Area},
  author =       {Liang, Jingsong and Wang, Zhichen and Cao, Yuhong and Chiun, Jimmy and Zhang, Mengqi and Sartoretti, Guillaume Adrien},
  booktitle = 	 {Proceedings of The 7th Conference on Robot Learning},
  pages = 	 {1425--1436},
  year = 	 {2023},
  editor = 	 {Tan, Jie and Toussaint, Marc and Darvish, Kourosh},
  volume = 	 {229},
  series = 	 {Proceedings of Machine Learning Research},
  month = 	 {06--09 Nov},
  publisher =    {PMLR},
  url = 	 {https://proceedings.mlr.press/v229/liang23a.html}
}

@article{Epstein2017TheCM,
  title={The cognitive map in humans: spatial navigation and beyond},
  author={Russell A. Epstein and Eva Zita Patai and Joshua B. Julian and Hugo J. Spiers},
  journal={Nature Neuroscience},
  year={2017},
  volume={20},
  pages={1504-1513},
  url={https://api.semanticscholar.org/CorpusID:25675}
}

@article{hurst2024gpt,
  title={Gpt-4o system card},
  author={Hurst, Aaron and Lerer, Adam and Goucher, Adam P and Perelman, Adam and Ramesh, Aditya and Clark, Aidan and Ostrow, AJ and Welihinda, Akila and Hayes, Alan and Radford, Alec and others},
  journal={arXiv preprint arXiv:2410.21276},
  year={2024}
}

@inproceedings{yang2025thinking,
  title={Thinking in space: How multimodal large language models see, remember, and recall spaces},
  author={Yang, Jihan and Yang, Shusheng and Gupta, Anjali W and Han, Rilyn and Fei-Fei, Li and Xie, Saining},
  booktitle={Proceedings of the Computer Vision and Pattern Recognition Conference},
  pages={10632--10643},
  year={2025}
}

@inproceedings{cheng2024navila,
title = {NaVILA: Legged Robot Vision-Language-Action Model for Navigation},
    author = {Cheng, An-Chieh and Ji, Yandong and Yang, Zhaojing and Zou, Xueyan and Kautz, Jan and Biyik, Erdem and Yin,
    Hongxu and Liu, Sifei and Wang, Xiaolong},
    booktitle = {RSS},
    year = {2025},
}

@inproceedings{zou20253d,
  title={3D-SPATIAL MULTIMODAL MEMORY},
  author={Zou, Xueyan and Song, Yuchen and Qiu, Ri-Zhao and Peng, Xuanbin and Ye, Jianglong and Liu, Sifei and Wang, Xiaolong},
  booktitle={The Thirteenth International Conference on Learning Representations},
  year={2025}
}

@article{yang2024think,
    title={{Thinking in Space: How Multimodal Large Language Models See, Remember and Recall Spaces}},
    author={Yang, Jihan and Yang, Shusheng and Gupta, Anjali and Han, Rilyn and Fei-Fei, Li and Xie, Saining},
    year={2024},
    journal={arXiv preprint arXiv:2412.14171},
}

@article{zhang2024uni,
  title={Uni-navid: A video-based vision-language-action model for unifying embodied navigation tasks},
  author={Zhang, Jiazhao and Wang, Kunyu and Wang, Shaoan and Li, Minghan and Liu, Haoran and Wei, Songlin and Wang, Zhongyuan and Zhang, Zhizheng and Wang, He},
  journal={arXiv preprint arXiv:2412.06224},
  year={2024}
}

@article{zhang2024navid,
  title={Navid: Video-based vlm plans the next step for vision-and-language navigation},
  author={Zhang, Jiazhao and Wang, Kunyu and Xu, Rongtao and Zhou, Gengze and Hong, Yicong and Fang, Xiaomeng and Wu, Qi and Zhang, Zhizheng and Wang, He},
  journal={arXiv preprint arXiv:2402.15852},
  year={2024}
}

@inproceedings{wang2025vggt,
  title={VGGT: Visual Geometry Grounded Transformer},
  author={Wang, Jianyuan and Chen, Minghao and Karaev, Nikita and Vedaldi, Andrea and Rupprecht, Christian and Novotny, David},
  booktitle={Proceedings of the IEEE/CVF Conference on Computer Vision and Pattern Recognition},
  year={2025}
}

@article{DBLP:journals/corr/abs-1711-11543,
  author       = {Abhishek Das and
                  Samyak Datta and
                  Georgia Gkioxari and
                  Stefan Lee and
                  Devi Parikh and
                  Dhruv Batra},
  title        = {Embodied Question Answering},
  journal      = {CoRR},
  volume       = {abs/1711.11543},
  year         = {2017},
  url          = {http://arxiv.org/abs/1711.11543},
  eprinttype    = {arXiv},
  eprint       = {1711.11543},
  bibsource    = {dblp computer science bibliography, https://dblp.org}
}

@article{DBLP:journals/corr/abs-1803-10122,
  author       = {David Ha and
                  J{\"{u}}rgen Schmidhuber},
  title        = {World Models},
  journal      = {CoRR},
  volume       = {abs/1803.10122},
  year         = {2018},
  url          = {http://arxiv.org/abs/1803.10122},
  eprinttype    = {arXiv},
  eprint       = {1803.10122},
  bibsource    = {dblp computer science bibliography, https://dblp.org}
}

@article{DBLP:journals/corr/abs-1912-11121,
  author       = {Alexander Sax and
                  Jeffrey O. Zhang and
                  Bradley Emi and
                  Amir Zamir and
                  Silvio Savarese and
                  Leonidas J. Guibas and
                  Jitendra Malik},
  title        = {Learning to Navigate Using Mid-Level Visual Priors},
  journal      = {CoRR},
  volume       = {abs/1912.11121},
  year         = {2019},
  url          = {http://arxiv.org/abs/1912.11121},
  eprinttype    = {arXiv},
  eprint       = {1912.11121},
  bibsource    = {dblp computer science bibliography, https://dblp.org}
}

@article{DBLP:journals/corr/abs-1901-03035,
  author       = {Chih{-}Yao Ma and
                  Jiasen Lu and
                  Zuxuan Wu and
                  Ghassan AlRegib and
                  Zsolt Kira and
                  Richard Socher and
                  Caiming Xiong},
  title        = {Self-Monitoring Navigation Agent via Auxiliary Progress Estimation},
  journal      = {CoRR},
  volume       = {abs/1901.03035},
  year         = {2019},
  url          = {http://arxiv.org/abs/1901.03035},
  eprinttype    = {arXiv},
  eprint       = {1901.03035},
  bibsource    = {dblp computer science bibliography, https://dblp.org}
}

@article{DBLP:journals/corr/abs-1905-12255,
  author       = {Vihan Jain and
                  Gabriel Magalh{\~{a}}es and
                  Alexander Ku and
                  Ashish Vaswani and
                  Eugene Ie and
                  Jason Baldridge},
  title        = {Stay on the Path: Instruction Fidelity in Vision-and-Language Navigation},
  journal      = {CoRR},
  volume       = {abs/1905.12255},
  year         = {2019},
  url          = {http://arxiv.org/abs/1905.12255},
  eprinttype    = {arXiv},
  eprint       = {1905.12255},
  bibsource    = {dblp computer science bibliography, https://dblp.org}
}

@article{DBLP:journals/corr/abs-2005-12256,
  author       = {Devendra Singh Chaplot and
                  Ruslan Salakhutdinov and
                  Abhinav Gupta and
                  Saurabh Gupta},
  title        = {Neural Topological {SLAM} for Visual Navigation},
  journal      = {CoRR},
  volume       = {abs/2005.12256},
  year         = {2020},
  url          = {https://arxiv.org/abs/2005.12256},
  eprinttype    = {arXiv},
  eprint       = {2005.12256},
  bibsource    = {dblp computer science bibliography, https://dblp.org}
}

@article{DBLP:journals/corr/abs-1711-00937,
  author       = {A{\"{a}}ron van den Oord and
                  Oriol Vinyals and
                  Koray Kavukcuoglu},
  title        = {Neural Discrete Representation Learning},
  journal      = {CoRR},
  volume       = {abs/1711.00937},
  year         = {2017},
  url          = {http://arxiv.org/abs/1711.00937},
  eprinttype    = {arXiv},
  eprint       = {1711.00937},
  bibsource    = {dblp computer science bibliography, https://dblp.org}
}

@misc{brohan2023rt2visionlanguageactionmodelstransfer,
      title={RT-2: Vision-Language-Action Models Transfer Web Knowledge to Robotic Control}, 
      author={Anthony Brohan and Noah Brown and Justice Carbajal and Yevgen Chebotar and Xi Chen and Krzysztof Choromanski and Tianli Ding and Danny Driess and Avinava Dubey and Chelsea Finn and Pete Florence and Chuyuan Fu and Montse Gonzalez Arenas and Keerthana Gopalakrishnan and Kehang Han and Karol Hausman and Alexander Herzog and Jasmine Hsu and Brian Ichter and Alex Irpan and Nikhil Joshi and Ryan Julian and Dmitry Kalashnikov and Yuheng Kuang and Isabel Leal and Lisa Lee and Tsang-Wei Edward Lee and Sergey Levine and Yao Lu and Henryk Michalewski and Igor Mordatch and Karl Pertsch and Kanishka Rao and Krista Reymann and Michael Ryoo and Grecia Salazar and Pannag Sanketi and Pierre Sermanet and Jaspiar Singh and Anikait Singh and Radu Soricut and Huong Tran and Vincent Vanhoucke and Quan Vuong and Ayzaan Wahid and Stefan Welker and Paul Wohlhart and Jialin Wu and Fei Xia and Ted Xiao and Peng Xu and Sichun Xu and Tianhe Yu and Brianna Zitkovich},
      year={2023},
      eprint={2307.15818},
      archivePrefix={arXiv},
      primaryClass={cs.RO},
      url={https://arxiv.org/abs/2307.15818}, 
}

@article{DBLP:journals/corr/abs-1807-03748,
  author       = {A{\"{a}}ron van den Oord and
                  Yazhe Li and
                  Oriol Vinyals},
  title        = {Representation Learning with Contrastive Predictive Coding},
  journal      = {CoRR},
  volume       = {abs/1807.03748},
  year         = {2018},
  url          = {http://arxiv.org/abs/1807.03748},
  eprinttype    = {arXiv},
  eprint       = {1807.03748},
  bibsource    = {dblp computer science bibliography, https://dblp.org}
}

@inproceedings{wang2024research,
  title={Research on autonomous robots navigation based on reinforcement learning},
  author={Wang, Zixiang and Yan, Hao and Wang, Yining and Xu, Zhengjia and Wang, Zhuoyue and Wu, Zhizhong},
  booktitle={2024 3rd International Conference on Robotics, Artificial Intelligence and Intelligent Control (RAIIC)},
  pages={78--81},
  year={2024},
  organization={IEEE}
}

@inproceedings{liu2024volumetric,
  title={Volumetric environment representation for vision-language navigation},
  author={Liu, Rui and Wang, Wenguan and Yang, Yi},
  booktitle={Proceedings of the IEEE/CVF conference on computer vision and pattern recognition},
  pages={16317--16328},
  year={2024}
}

@inproceedings{zheng2024towards,
  title={Towards learning a generalist model for embodied navigation},
  author={Zheng, Duo and Huang, Shijia and Zhao, Lin and Zhong, Yiwu and Wang, Liwei},
  booktitle={Proceedings of the IEEE/CVF Conference on Computer Vision and Pattern Recognition},
  pages={13624--13634},
  year={2024}
}

@inproceedings{majumdar2024openeqa,
  title={Openeqa: Embodied question answering in the era of foundation models},
  author={Majumdar, Arjun and Ajay, Anurag and Zhang, Xiaohan and Putta, Pranav and Yenamandra, Sriram and Henaff, Mikael and Silwal, Sneha and Mcvay, Paul and Maksymets, Oleksandr and Arnaud, Sergio and others},
  booktitle={Proceedings of the IEEE/CVF conference on computer vision and pattern recognition},
  pages={16488--16498},
  year={2024}
}

@inproceedings{yu2025seqafford,
  title={Seqafford: Sequential 3d affordance reasoning via multimodal large language model},
  author={Yu, Chunlin and Wang, Hanqing and Shi, Ye and Luo, Haoyang and Yang, Sibei and Yu, Jingyi and Wang, Jingya},
  booktitle={Proceedings of the Computer Vision and Pattern Recognition Conference},
  pages={1691--1701},
  year={2025}
}

@article{mu2023embodiedgpt,
  title={Embodiedgpt: Vision-language pre-training via embodied chain of thought},
  author={Mu, Yao and Zhang, Qinglong and Hu, Mengkang and Wang, Wenhai and Ding, Mingyu and Jin, Jun and Wang, Bin and Dai, Jifeng and Qiao, Yu and Luo, Ping},
  journal={Advances in Neural Information Processing Systems},
  volume={36},
  pages={25081--25094},
  year={2023},
}

@article{ji2024neds,
  title={Neds-slam: A neural explicit dense semantic slam framework using 3d gaussian splatting},
  author={Ji, Yiming and Liu, Yang and Xie, Guanghu and Ma, Boyu and Xie, Zongwu and Liu, Hong},
  journal={IEEE Robotics and Automation Letters},
  year={2024},
  publisher={IEEE}
}

@inproceedings{sun2024high,
  title={High-fidelity slam using gaussian splatting with rendering-guided densification and regularized optimization},
  author={Sun, Shuo and Mielle, Malcolm and Lilienthal, Achim J and Magnusson, Martin},
  booktitle={2024 IEEE/RSJ International Conference on Intelligent Robots and Systems (IROS)},
  pages={10476--10482},
  year={2024},
  organization={IEEE}
}

@inproceedings{octo_2023,
    title={Octo: An Open-Source Generalist Robot Policy},
    author = {{Octo Model Team} and Dibya Ghosh and Homer Walke and Karl Pertsch and Kevin Black and Oier Mees and Sudeep Dasari and Joey Hejna and Charles Xu and Jianlan Luo and Tobias Kreiman and {You Liang} Tan and Lawrence Yunliang Chen and Pannag Sanketi and Quan Vuong and Ted Xiao and Dorsa Sadigh and Chelsea Finn and Sergey Levine},
    booktitle = {Proceedings of Robotics: Science and Systems},
    address  = {Delft, Netherlands},
    year = {2024},
}

@inproceedings{10.1609/aaai.v38i7.28597,
author = {Zhou, Gengze and Hong, Yicong and Wu, Qi},
title = {NavGPT: explicit reasoning in vision-and-language navigation with large language models},
year = {2024},
isbn = {978-1-57735-887-9},
publisher = {AAAI Press},
url = {https://doi.org/10.1609/aaai.v38i7.28597},
doi = {10.1609/aaai.v38i7.28597},
articleno = {849},
numpages = {9},
series = {AAAI'24/IAAI'24/EAAI'24}
}

@article{zhu2023ponderv2,
  title={PonderV2: Pave the Way for 3D Foundation Model with A Universal Pre-training Paradigm}, 
  author={Haoyi Zhu and Honghui Yang and Xiaoyang Wu and Di Huang and Sha Zhang and Xianglong He and Tong He and Hengshuang Zhao and Chunhua Shen and Yu Qiao and Wanli Ouyang},
  journal={arXiv preprint arXiv:2310.08586},
  year={2023}
}

@inproceedings{huang2023ponder,
  title={Ponder: Point cloud pre-training via neural rendering},
  author={Huang, Di and Peng, Sida and He, Tong and Yang, Honghui and Zhou, Xiaowei and Ouyang, Wanli},
  booktitle={Proceedings of the IEEE/CVF International Conference on Computer Vision},
  pages={16089--16098},
  year={2023}
}

@article{yang2023unipad,
  title={UniPAD: A Universal Pre-training Paradigm for Autonomous Driving}, 
  author={Honghui Yang and Sha Zhang and Di Huang and Xiaoyang Wu and Haoyi Zhu and Tong He and Shixiang Tang and Hengshuang Zhao and Qibo Qiu and Binbin Lin and Xiaofei He and Wanli Ouyang},
  journal={arXiv preprint arXiv:2310.08370},
  year={2023},
}

@inproceedings{10.5555/3618408.3618748,
author = {Driess, Danny and Xia, Fei and Sajjadi, Mehdi S. M. and Lynch, Corey and Chowdhery, Aakanksha and Ichter, Brian and Wahid, Ayzaan and Tompson, Jonathan and Vuong, Quan and Yu, Tianhe and Huang, Wenlong and Chebotar, Yevgen and Sermanet, Pierre and Duckworth, Daniel and Levine, Sergey and Vanhoucke, Vincent and Hausman, Karol and Toussaint, Marc and Greff, Klaus and Zeng, Andy and Mordatch, Igor and Florence, Pete},
title = {PaLM-E: an embodied multimodal language model},
year = {2023},
publisher = {JMLR.org},
booktitle = {Proceedings of the 40th International Conference on Machine Learning},
articleno = {340},
numpages = {20},
location = {Honolulu, Hawaii, USA},
series = {ICML'23}
}

@inproceedings{wang2025causality,
  title={Causality-Aware Transformer Networks for Robotic Navigation},
  author={Wang, Ruoyu and Liu, Yao and Cao, Yuanjiang and Yao, Lina},
  booktitle={International Conference on Neural Information Processing},
  pages={403--418},
  year={2025},
  organization={Springer}
}

@article{jain2023mnemosyne,
  title={Mnemosyne: Learning to train transformers with transformers},
  author={Jain, Deepali and Choromanski, Krzysztof M and Dubey, Kumar Avinava and Singh, Sumeet and Sindhwani, Vikas and Zhang, Tingnan and Tan, Jie},
  journal={Advances in Neural Information Processing Systems},
  volume={36},
  pages={77331--77358},
  year={2023}
}

@article{wu2025spatial,
  title={Spatial-mllm: Boosting mllm capabilities in visual-based spatial intelligence},
  author={Wu, Diankun and Liu, Fangfu and Hung, Yi-Hsin and Duan, Yueqi},
  journal={arXiv preprint arXiv:2505.23747},
  year={2025}
}

@inproceedings{huang2021spatio,
  title={Spatio-temporal self-supervised representation learning for 3d point clouds},
  author={Huang, Siyuan and Xie, Yichen and Zhu, Song-Chun and Zhu, Yixin},
  booktitle={Proceedings of the IEEE/CVF international conference on computer vision},
  pages={6535--6545},
  year={2021}
}

@inproceedings{zhao2023generative,
  title={Generative causal interpretation model for spatio-temporal representation learning},
  author={Zhao, Yu and Deng, Pan and Liu, Junting and Jia, Xiaofeng and Zhang, Jianwei},
  booktitle={Proceedings of the 29th ACM SIGKDD Conference on Knowledge Discovery and Data Mining},
  pages={3537--3548},
  year={2023}
}

@inproceedings{zhang2024multi,
  title={Multi-scale video anomaly detection by multi-grained spatio-temporal representation learning},
  author={Zhang, Menghao and Wang, Jingyu and Qi, Qi and Sun, Haifeng and Zhuang, Zirui and Ren, Pengfei and Ma, Ruilong and Liao, Jianxin},
  booktitle={Proceedings of the IEEE/CVF Conference on Computer Vision and Pattern Recognition},
  pages={17385--17394},
  year={2024}
}

@article{qi2021self,
  title={Self-regulated learning for egocentric video activity anticipation},
  author={Qi, Zhaobo and Wang, Shuhui and Su, Chi and Su, Li and Huang, Qingming and Tian, Qi},
  journal={IEEE transactions on pattern analysis and machine intelligence},
  volume={45},
  number={6},
  pages={6715--6730},
  year={2021},
  publisher={IEEE}
}

@inproceedings{akiva2023self,
  title={Self-supervised object detection from egocentric videos},
  author={Akiva, Peri and Huang, Jing and Liang, Kevin J and Kovvuri, Rama and Chen, Xingyu and Feiszli, Matt and Dana, Kristin and Hassner, Tal},
  booktitle={Proceedings of the IEEE/CVF International Conference on Computer Vision},
  pages={5225--5237},
  year={2023}
}

@inproceedings{planamente2021self,
  title={Self-supervised joint encoding of motion and appearance for first person action recognition},
  author={Planamente, Mirco and Bottino, Andrea and Caputo, Barbara},
  booktitle={2020 25th International Conference on Pattern Recognition (ICPR)},
  pages={8751--8758},
  year={2021},
  organization={IEEE}
}

@article{xiang2023language,
  title={Language models meet world models: Embodied experiences enhance language models},
  author={Xiang, Jiannan and Tao, Tianhua and Gu, Yi and Shu, Tianmin and Wang, Zirui and Yang, Zichao and Hu, Zhiting},
  journal={Advances in neural information processing systems},
  volume={36},
  pages={75392--75412},
  year={2023}
}

@inproceedings{zhou2024navgpt,
  title={Navgpt: Explicit reasoning in vision-and-language navigation with large language models},
  author={Zhou, Gengze and Hong, Yicong and Wu, Qi},
  booktitle={Proceedings of the AAAI Conference on Artificial Intelligence},
  volume={38},
  number={7},
  pages={7641--7649},
  year={2024}
}

@inproceedings{zhou2024navgpt2,
  title={Navgpt-2: Unleashing navigational reasoning capability for large vision-language models},
  author={Zhou, Gengze and Hong, Yicong and Wang, Zun and Wang, Xin Eric and Wu, Qi},
  booktitle={European Conference on Computer Vision},
  pages={260--278},
  year={2024},
  organization={Springer}
}

@inproceedings{DBLP:conf/rss/ZhangWXZHF0ZW24,
  author={Jiazhao Zhang and Kunyu Wang and Rongtao Xu and Gengze Zhou and Yicong Hong and Xiaomeng Fang and Qi Wu and Zhizheng Zhang and He Wang},
  title={NaVid: Video-based VLM Plans the Next Step for Vision-and-Language Navigation},
  year={2024},
  cdate={1704067200000},
  url={https://doi.org/10.15607/RSS.2024.XX.079},
  booktitle={Robotics: Science and Systems}
}

@article{wei2022chain,
  title={Chain-of-thought prompting elicits reasoning in large language models},
  author={Wei, Jason and Wang, Xuezhi and Schuurmans, Dale and Bosma, Maarten and Xia, Fei and Chi, Ed and Le, Quoc V and Zhou, Denny and others},
  journal={Advances in neural information processing systems},
  volume={35},
  pages={24824--24837},
  year={2022}
}

@inproceedings{krantz_vlnce_2020,
  title={Beyond the Nav-Graph: Vision and Language Navigation in Continuous Environments},
  author={Jacob Krantz and Erik Wijmans and Arjun Majundar and Dhruv Batra and Stefan Lee},
  booktitle={European Conference on Computer Vision (ECCV)},
  year={2020}
 }

@article{berg2025semantic,
  title={Semantic World Models},
  author={Berg, Jacob and Zhu, Chuning and Bao, Yanda and Durugkar, Ishan and Gupta, Abhishek},
  journal={arXiv preprint arXiv:2510.19818},
  year={2025}
}

@inproceedings{
wang2025remi,
title={{REMI}: Reconstructing Episodic Memory During Internally Driven Path Planning},
author={Zhaoze Wang and Genela Morris and Dori Derdikman and Pratik Chaudhari and Vijay Balasubramanian},
booktitle={The Thirty-ninth Annual Conference on Neural Information Processing Systems},
year={2025},
url={https://openreview.net/forum?id=LPWzV8zrgj}
}

@article{lin2025navcot,
  title={Navcot: Boosting llm-based vision-and-language navigation via learning disentangled reasoning},
  author={Lin, Bingqian and Nie, Yunshuang and Wei, Ziming and Chen, Jiaqi and Ma, Shikui and Han, Jianhua and Xu, Hang and Chang, Xiaojun and Liang, Xiaodan},
  journal={IEEE Transactions on Pattern Analysis and Machine Intelligence},
  year={2025},
  publisher={IEEE}
}

@inproceedings{chenlongvila,
  title={LongVILA: Scaling Long-Context Visual Language Models for Long Videos},
  author={Chen, Yukang and Xue, Fuzhao and Li, Dacheng and Hu, Qinghao and Zhu, Ligeng and Li, Xiuyu and Fang, Yunhao and Tang, Haotian and Yang, Shang and Liu, Zhijian and others},
  booktitle={The Thirteenth International Conference on Learning Representations}
}

@article{yang2024qwen2,
  title={Qwen2 Technical Report},
  author={Yang, An and Yang, Baosong and Hui, Binyuan and Zheng, Bo and Yu, Bowen and Zhou, Chang and Li, Chengpeng and Li, Chengyuan and Liu, Dayiheng and Huang, Fei and others},
  journal={CoRR},
  year={2024}
}

@article{tschannen2025siglip,
  title={Siglip 2: Multilingual vision-language encoders with improved semantic understanding, localization, and dense features},
  author={Tschannen, Michael and Gritsenko, Alexey and Wang, Xiao and Naeem, Muhammad Ferjad and Alabdulmohsin, Ibrahim and Parthasarathy, Nikhil and Evans, Talfan and Beyer, Lucas and Xia, Ye and Mustafa, Basil and others},
  journal={arXiv preprint arXiv:2502.14786},
  year={2025}
}

\clearpage
\setcounter{page}{1}
\maketitlesupplementary

\section{Model and Training Details}
\subsection{Train Data}
\label{supp:data}
To accomplish the dual tasks of visual navigation and spatial reasoning, our training dataset is constructed directly upon the \textbf{VLN-CE (Vision-and-Language Navigation in Continuous Environments) training set}.

Our core training curriculum is the \textbf{MindCraft-Train dataset}. As detailed in the main paper (Section 3.3), this dataset is generated by our procedural pipeline, which traverses the expert trajectories provided by VLN-CE within the Matterport3D environments. For each trajectory, our pipeline injects the three types of spatio-temporal cognitive queries (retrospective, introspective, and prospective) and logs the ground-truth answers.

This approach ensures that the agent is trained on a unified dataset where navigation actions and reasoning queries are concurrent. All models, including baselines adapted for our task, were trained exclusively on this dataset, ensuring a fair comparison. This single-source training setup is critical for validating our hypothesis that the reasoning task acts as an effective auxiliary supervision for improving navigation, and allows for a true \textbf{zero-shot} evaluation on downstream tasks like VSI-Bench, as neither the environments (for val-unseen splits) nor the general VQA task formats were seen during training.

\begin{figure}[ht!]
    \centering
    \includegraphics[width=1\linewidth]{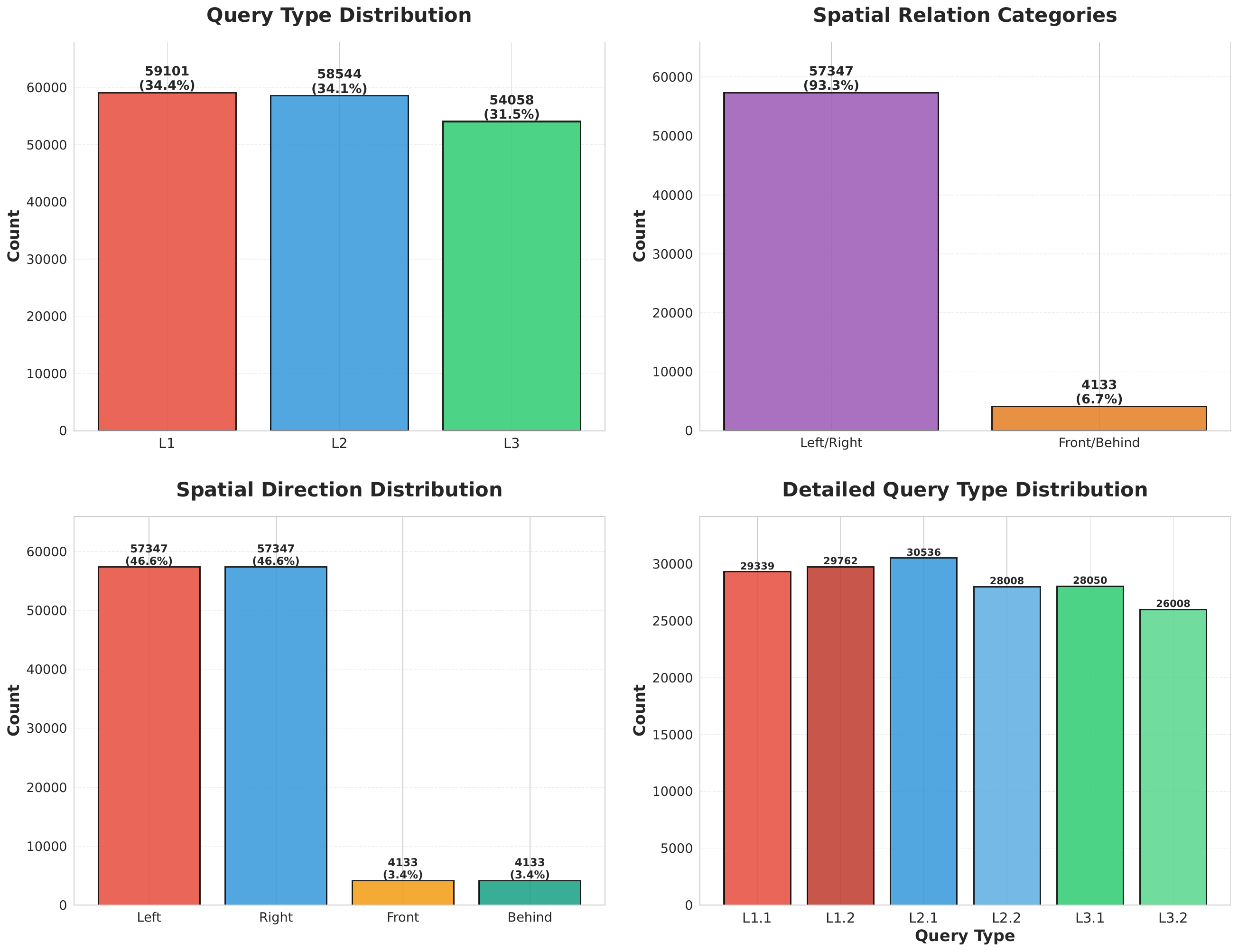}
    \caption{\textbf{MindCraft Data Statistics.}}
    \label{fig:placeholder}
\end{figure}

\subsection{MindCraft Dataset Generation}
\label{app:dataset}

For each expert trajectory, our pipeline programmatically generates a corresponding MindCraft dataset through a structured, three-stage process designed to ensure that all generated queries are contextually relevant, unambiguous, and grounded in the agent's experiential history.

First, during Trajectory Traversal and Logging, a virtual agent precisely follows the expert path. At each step, it constructs a ``Path Memory Log," a comprehensive record of its sensory experience. This log systematically captures the set of visible semantic objects, their properties, and the agent's spatiotemporal state, effectively chronicling what was perceived, where, and when.

Second, in the Cognitive Queries Generation stage, the pipeline synthesizes a query that targets a specific facet of spatial cognition. Our framework defines six query categories across three cognitive tiers: L1) Memory Recall (Retrospective Queries), L2) Situational Awareness (Introspective Queries), and L3) Predictive Reasoning (Prospective Queries). Crucially, the instantiation of a query is not arbitrary but is instead governed by a set of strict, task-specific preconditions. The pipeline first assesses the trajectory log to determine which types of cognitive queries are valid and meaningful to generate for that specific path. For instance, an Object Attribute Recall query is contingent upon the agent having previously observed an object that meets two criteria: (1) its bounding box occupied over 10\% of the viewport area at the time of observation, and it is no longer in the current field of view. (2) To ensure referential unambiguity, the pipeline validates against the simulator's ground-truth scene graph that no other instance of the same semantic category (e.g., another ``chair") is present within a 5-meter radius of the original observation's location.

Only the query types whose preconditions are satisfied by the agent's memory log become candidates for generation. From this valid set, one is selected and formulated.

Finally, during Ground-Truth Answer Generation, the pipeline programmatically derives the correct answer by leveraging the simulator's ground-truth information. The answer is computed based on the veridical state of the 3D environment and the agent's logged history. For example, the spatial relationship between the agent and a recalled object is determined by its position within the agent's egocentric visual field at the moment of initial observation.

\subsection{Train Strategy}
\label{supp:strategy}
We train the LASAR model end-to-end using the composite objective function described in the main paper. The model is optimized to jointly predict the navigation action $a_t$ (via imitation learning) and, when present, answer the cognitive query $ans_t$ (via sequence generation). To enhance the model's performance on video spatial understanding tasks, we fine-tuned the Vicuna-7B model using LoRA (Low-Rank Adaptation). The core idea of LoRA is to achieve parameter-efficient optimization by inserting trainable low-rank matrices without making large-scale adjustments to the original weights of the pre-trained model. This approach allows the model to adapt to downstream task requirements while maintaining its performance.

Specifically, LoRA decomposes the weight matrix $W \in \mathbb{R}^{d \times k}$ into two low-rank matrices $A \in \mathbb{R}^{d \times r}$ and $B \in \mathbb{R}^{r \times k}$. Through low-rank decomposition, the weight update formula can be expressed as $W' = W + A \cdot B$. During this process, the original weights $W$ remain frozen, and only the newly added low-rank matrices $A$ and $B$ are optimized. We inserted LoRA modules into key components of the model, including the query projection (\texttt{q\_proj}), value projection (\texttt{v\_proj}), and language model head (\texttt{lm\_head}), to enhance the model's ability to model dynamic relationships between video frames and perform cross-modal reasoning.

The full loss $\mathcal{L}_{\text{total}}$ combines the main task loss $\mathcal{L}_{\text{MindCraft}}$ with the three crucial auxiliary losses: $\mathcal{L}_{\text{crl}}$ (Spatio-temporal Contextual Representation Learning), $\mathcal{L}_{\text{sem}}$ (Semantic Atlas Learning), and $\mathcal{L}_{\text{retro}}$ (Episodic Discriminability).

For video spatial understanding tasks, the use of LoRA allows us to efficiently optimize the model, making it more adaptable to downstream tasks while maintaining reasonable computational resource usage. This fine-tuning approach provides strong technical support for our video spatial understanding tasks.

\subsection{Hyperparameter Details}
\label{app:hyperparams}

To ensure the reproducibility of our results, we provide a comprehensive list of the key hyperparameters used for training the LASAR model. We utilized the AdamW optimizer for stable and efficient training. The primary hyperparameters for the optimization and training process are summarized in Table~\ref{tab:hyperparams}. These settings were kept consistent across all main experiments unless otherwise specified in the ablation studies.

\begin{table}[ht]
    \centering
    \caption{Key hyperparameters for training the LASAR framework.}
    \label{tab:hyperparams}
    \renewcommand{\arraystretch}{1.2} 
    \resizebox{\linewidth}{!}{%
        \begin{tabular}{@{}ll@{}}
            \toprule
            \textbf{Hyperparameter} & \textbf{Value} \\
            \midrule
            \multicolumn{2}{l}{\textit{Optimizer \& Scheduler}} \\
            \quad Optimizer & AdamW \\
            \quad Learning Rate (Peak) & $1 \times 10^{-4}$ \\
            \quad Betas ($\beta_1, \beta_2$) & (0.9, 0.999) \\
            \quad Weight Decay & 0.01 \\
            \quad Learning Rate Schedule & Cosine decay \\
            \quad Warmup Epochs & 1 \\
            \midrule
            \multicolumn{2}{l}{\textit{Training \& Batching}} \\
            \quad Total Training Epochs & 10 \\
            \quad Per-Device Batch Size & 8 \\
            \quad Gradient Accumulation Steps & 8 \\
            \quad Effective Batch Size & 64 ($8 \times 8 \text{ GPUs}$) \\
            \quad Mixed Precision & bfloat16 \\
            \midrule
            \multicolumn{2}{l}{\textit{LoRA Fine-tuning}} \\
            \quad LoRA Rank ($r$) & 16 \\
            \quad LoRA Alpha ($\alpha$) & 32 \\
            \quad LoRA Dropout & 0.1 \\
            \quad LoRA Target Modules & \texttt{q\_proj}, \texttt{v\_proj}, \texttt{lm\_head} \\
            \midrule
            \multicolumn{2}{l}{\textit{Loss Configuration}} \\
            \quad Query Answer Weight ($\lambda_{qa}$) & 1.0 \\ 
            \quad ST-CRL Loss Weight ($\lambda_{c}$) & 0.1 \\ 
            \quad Semantic Atlas Loss Weight ($\lambda_{s}$) & 0.2 \\ 
            \quad Episodic Loss Weight ($\lambda_{r}$) & 0.1 \\ 
            \quad Contrastive Temperature ($\tau$) & 0.07 \\ 
            \quad ST-CRL Negative Samples (N) & 64 \\ 
            \bottomrule
        \end{tabular}%
    }
\end{table}

\paragraph{LoRA Configuration.}
As detailed in the `Train Strategy` subsection, we employed Low-Rank Adaptation (LoRA) for parameter-efficient fine-tuning. The rank ($r$) was set to 16, providing a good balance between expressiveness and parameter efficiency. The scaling factor ($\alpha$) was set to 32. To prevent overfitting on the adapter weights, a dropout rate of 0.1 was applied specifically to the LoRA modules. We targeted the query (\texttt{q\_proj}) and value (\texttt{v\_proj}) projections within the VLM's self-attention layers, as well as the final language model head (\texttt{lm\_head}), as these are critical for adapting the model's reasoning and generation capabilities to our specific tasks.

\section{Model Architecture Details}
\label{app:architecture_details}

Our proposed framework, LASAR, is constructed around a central Vision-Language Model (VLM) and is augmented by a series of specialized encoders. Each encoder is designed to process a specific modality (vision, geometry, trajectory), transforming raw sensory input into rich feature representations. These features are then projected into a common embedding space, allowing the core VLM to perform holistic reasoning across all available information streams. Below, we detail the specifics of each key component.

\paragraph{Core VLM Backbone.}
The heart of our model is the \textbf{Vicuna-7B (v1.5)}, a powerful pretrained VLM that serves as our primary reasoning engine and Unified Decision Head. We leverage its advanced capabilities in understanding and integrating vision and language. The input to this model is a carefully structured sequence of embeddings from the episodic memory and the spatial semantic memory, as described in the main paper.

\paragraph{Visual Semantic Encoder.}
To extract high-level semantic information from each RGB frame $I_t$, we employ a frozen pretrained \textbf{Siglip} model. For each frame, we extract the final feature representation $F_{\text{vis}, t}$, which captures the objects and their appearance. These features are then passed through a projection layer to match the VLM's hidden dimension, forming the input for the Geometric-Semantic Fusion module.

\paragraph{Image-based Geometric Encoder.}
Complementary to the semantic features, we extract immediate, view-dependent geometric cues directly from the 2D image $I_t$. For this, we use a specialized visual-geometry encoder, specifically a version of \textbf{VGGT}~\cite{wang2025vggt} pretrained on multiple geometry estimation tasks (e.g., point cloud prediction from 2D). This model processes the RGB frame and outputs a dense feature map that implicitly encodes geometric information $F_{\text{geo}, t}$. These geometric features provide the spatial context necessary to ground the semantic features $F_{\text{vis}, t}$ through our cross-attention mechanism, as defined in the main paper (Equation 2).

\section{Evaluation Protocol Details}
\label{supp:metric}

\subsection{Overall Reasoning Accuracy (QA-Acc)}

This metric measures the overall accuracy of the agent's answers to all cognitive queries in the test set. Let $\mathcal{Q}$ be the set of all queries in the test set. For any query $q \in \mathcal{Q}$, let $A_q$ be the agent's answer and $G_q$ be the ground-truth answer. Let $\mathbb{I}(\cdot)$ be the indicator function.

The Overall Reasoning Accuracy (QA-Acc) is calculated as:
$$
\text{QA-Acc} = \frac{1}{|\mathcal{Q}|} \sum_{q \in \mathcal{Q}} \mathbb{I}(A_q = G_q)
$$

\subsection{Goal-Conditioned Accuracy (GCA)}

This metric measures the query accuracy \emph{only} on trajectories where the agent completed its navigation task (i.e., Success Rate $SR=1$). Let $\mathcal{T}_{succ}$ be the set of all trajectories where navigation was successful.
Let $\mathcal{Q}_{succ} \subseteq \mathcal{Q}$ be the subset of all queries that were posed during the trajectories in $\mathcal{T}_{succ}$.

The Goal-Conditioned Accuracy (GCA) is calculated as:
$$
\text{GCA} = \frac{1}{|\mathcal{Q}_{succ}|} \sum_{q \in \mathcal{Q}_{succ}} \mathbb{I}(A_q = G_q)
$$

\subsection{Cognitive Map Consistency (CMC)}

This metric measures the stability of the agent's internal cognitive map. This metric calculates the consistency of the agent's answers to \textbf{Equivalent Probe Sets}. An ``Equivalent Probe Set" $S_i$ contains $k_i$ queries $\{q_1, \dots, q_{k_i}\}$ that all target the \textbf{same underlying spatial fact} and are designed to have the \textbf{exact same ground-truth answer}.

The sources for constructing these sets $S_i$ include:
\begin{enumerate}[label=\textbf{\arabic*.}]
    \item \textbf{Template \& Relational Equivalence:} The query's phrasing or logical frame of reference is different, but the semantics and ground-truth answer are identical.
    \textit{(e.g., ``Is the sofa to the left of the lamp?" vs. ``Is the lamp to the right of the sofa?" GT is ``Yes" for both)}
    
    \item \textbf{Temporal Misalignment:} Queries within the same trajectory that are based on temporally shifted clips (e.g., offset by 1-2 steps), but the core fact and GT answer remain unchanged.
    
    \item \textbf{Cross-Query-Type Equivalence:} Targeting the same fact using queries of different cognitive types (e.g., introspective, retrospective).
    \textit{(e.g., An \textbf{introspective} query at $t=10$: ``What color is the vase you see now?" (GT: ``Red") vs. a \textbf{retrospective} query at $t=25$: ``What color was the vase you passed earlier?" (GT: ``Red"))}
    
    \item \textbf{Cross-Trajectory Metadata:} Queries from different trajectories that are determined (via metadata like environment/object ID) to be asking about the same fact with the same GT answer.
\end{enumerate}

We partition all queries into $M$ such Equivalent Probe Sets $\mathcal{S} = \{S_1, S_2, ..., S_M\}$. We only consider sets where $k_i \ge 2$.

\begin{enumerate}
    \item \textbf{Calculate Total Consistent Pairs ($C_{total}$):}
    We sum the number of pairs of queries within each set $S_i$ that received an identical answer ($A_{q_m} = A_{q_n}$).
    $$
    C_{total} = \sum_{i=1}^{M} \left( \sum_{m=1}^{k_i-1} \sum_{n=m+1}^{k_i} \mathbb{I}(A_{q_m} = A_{q_n}) \right)
    $$

    \item \textbf{Calculate Total Pairwise Comparisons ($P_{total}$):}
    We sum the total number of unique pairwise comparisons possible within each set $S_i$. $\binom{k_i}{2}$ is the binomial coefficient ``k choose 2".
    $$
    P_{total} = \sum_{i=1}^{M} \binom{k_i}{2} = \sum_{i=1}^{M} \frac{k_i (k_i - 1)}{2}
    $$

    \item \textbf{Calculate Final CMC:}
    The CMC is the ratio of total consistent pairs to the total pairwise comparisons.
    $$
    \text{CMC} = \frac{C_{total}}{P_{total}}
    $$
\end{enumerate}

\subsection{Reasoning Failure Impact (SR@WA)}

This metric measures the Navigation Success Rate \emph{only} on trajectories where the agent answered \textbf{at least one query incorrectly}. Let $\mathcal{T}$ be the set of all trajectories.
Let $\mathcal{T}_{WA} \subseteq \mathcal{T}$ (Wrong Answer) be the subset of trajectories containing at least one incorrect answer.
For any trajectory $t \in \mathcal{T}$, let $S(t) \in \{0, 1\}$ be its navigation success status (1 = success, 0 = failure).

The SR@WA is calculated as:
$$
\text{SR@WA} = \frac{1}{|\mathcal{T}_{WA}|} \sum_{t \in \mathcal{T}_{WA}} S(t)
$$

\subsection{Cross-dataset Evaluation}

\paragraph{Metrics Calculation.} All metrics were calculated using the official evaluation scripts provided by the respective benchmarks. For VLN-CE, we report Success Rate (SR), Success rate weighted by Path Length (SPL), \textbf{and for the R2R split, Oracle Success (OS)}. For VSI-Bench, we report Accuracy (ACC) for multiple-choice questions and Mean Relative Accuracy (MRA) for numerical questions.
\paragraph{Inference Setup.} All our models were evaluated on a single NVIDIA A100 GPU. We used greedy decoding (i.e., beam size of 1) for generating all responses to ensure efficiency and deterministic outputs. All reported scores are the average of three evaluation runs with different random seeds to ensure statistical stability.

\section{Additional Experimental Results}

\subsection{Baseline details}
\label{app:benchmark_details} 

\paragraph{Specialized Navigation Models}
For all specialized navigation baselines on VLN-CE, including \texttt{Navid}, \texttt{NaviLLM}, and \texttt{NaVILA}, we adhered to the following protocol to ensure a fair and reproducible comparison:
\begin{itemize}
    \item \textbf{Source:} We utilized the official codebases and pre-trained model weights released by the respective authors. No architectural modifications were made.
    \item \textbf{Evaluation Protocol:} We followed the standard evaluation scripts and environment setups provided with each baseline's repository. Results were generated on the \texttt{val-unseen} split as reported in the original papers.
\end{itemize}

\paragraph{General Vision-Language Models}
For all general-purpose VLMs evaluated on VSI-Bench, such as \texttt{GPT-4o}, \texttt{Gemini-1.5 Pro}, the \texttt{LLaVA} series, and the \texttt{Qwen} series, we used the following zero-shot evaluation setup:
\begin{itemize}
    \item \textbf{Model Version:} We used the latest available official APIs for proprietary models (e.g., \texttt{gpt-4o-2024-05-13}) and the official Hugging Face implementations for open-source models.
    \item \textbf{Prompting Strategy:} A consistent, minimal prompt template was employed across all models to query their spatial reasoning capabilities without providing few-shot examples. The templates were structured as follows:

    \begin{quote}
    \small
    \textbf{For Multiple-Choice Questions:}\\
    \texttt{The following is a video of an indoor scene. Based on the video, answer the following question by choosing the best option.
    <video\_placeholder>
    Question: [Question from VSI-Bench]
    Options:
    (A) [Option A]
    (B) [Option B]
    (C) [Option C]
    (D) [Option D]
    Answer (Provide the letter only):}
    \end{quote}
    
    \begin{quote}
    \small
    \textbf{For Numerical Questions:}\\
    \texttt{The following is a video of an indoor scene. Based on the video, answer the following question. Provide only the numerical value in your answer.
    <video\_placeholder>
    Question: [Question from VSI-Bench]
    Answer:}
    \end{quote}
\end{itemize}

\subsection{Computational Cost and Efficiency}
\label{supp:cost}

\paragraph{Training Efficiency}
Our training framework is designed to be resource-efficient despite the complexity of the multi-modal architecture. By leveraging the frozen pre-trained representations from Siglip and VGGT, and employing Low-Rank Adaptation (LoRA) for the Vicuna-7B backbone, we significantly reduce the computational burden. 
Specifically, with a LoRA rank of $r=16$, the number of trainable parameters is approximately \textbf{160M}, constituting less than \textbf{2\%} of the total model size ($\sim$\textbf{8.4B} parameters).
Training was conducted on 8 NVIDIA A100 (80GB) GPUs. The entire training process on the MindCraft-Train dataset required approximately \textbf{6 days}. We utilized mixed-precision training (bfloat16) and a global batch size of 64, with a peak VRAM usage of approximately \textbf{60GB} per GPU during training.

\paragraph{Inference Latency and Memory Footprint}
During inference, we evaluate the model on a single NVIDIA A100 GPU. The inference process involves three main stages: visual and geometric feature extraction, latent cognitive map querying, and LLM token generation.
\begin{itemize}
    \item \textbf{Latency:} The average inference latency per step is approximately \textbf{0.85s}. Compared to the navigation-only baseline (e.g., NaVILA), our model introduces a marginal overhead of only $\sim$\textbf{0.03s}. This slight increase is negligible in practice, especially considering the significant gains in reasoning accuracy (+4.7\% QA-Acc) and navigation success.
    \item \textbf{Memory:} The peak VRAM usage during inference is approximately \textbf{22GB}. This memory footprint is well within the capacity of high-end consumer-grade GPUs (e.g., NVIDIA RTX 3090/4090), making our model feasible for deployment on workstations. While the Episodic Memory grows with trajectory length, our frame sampling strategy ($K=32$) ensures that memory consumption remains bounded and predictable.
\end{itemize}

\paragraph{Summary}
In conclusion, LASAR achieves a superior balance between performance and efficiency. The inclusion of the Latent Cognitive Map and geometric encoders provides substantial improvements in spatial reasoning capabilities with minimal cost to inference speed and a manageable memory footprint.

\subsection{Additional Qualitative Results}
\label{supp:addqual}

\begin{figure}[ht!]
    \centering
    \includegraphics[width=\linewidth]{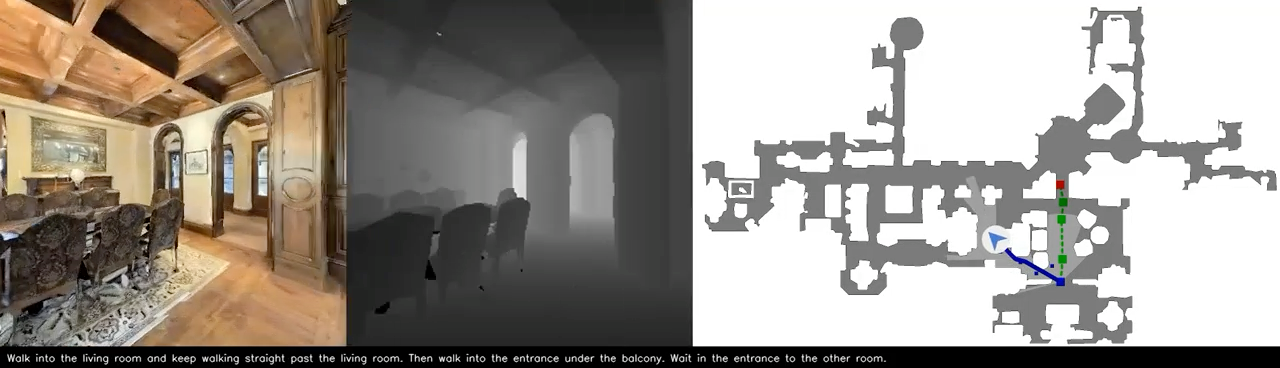}
    \caption{\textbf{Navigation Failure Case.} The model fails to execute the instruction ``Walk into the living room and keep walking straight past the living room..." due to a misinterpretation of the global room layout.}
    \label{fig:nav_failure}
\end{figure}

\begin{figure}[ht!]
    \centering
    \includegraphics[width=\linewidth]{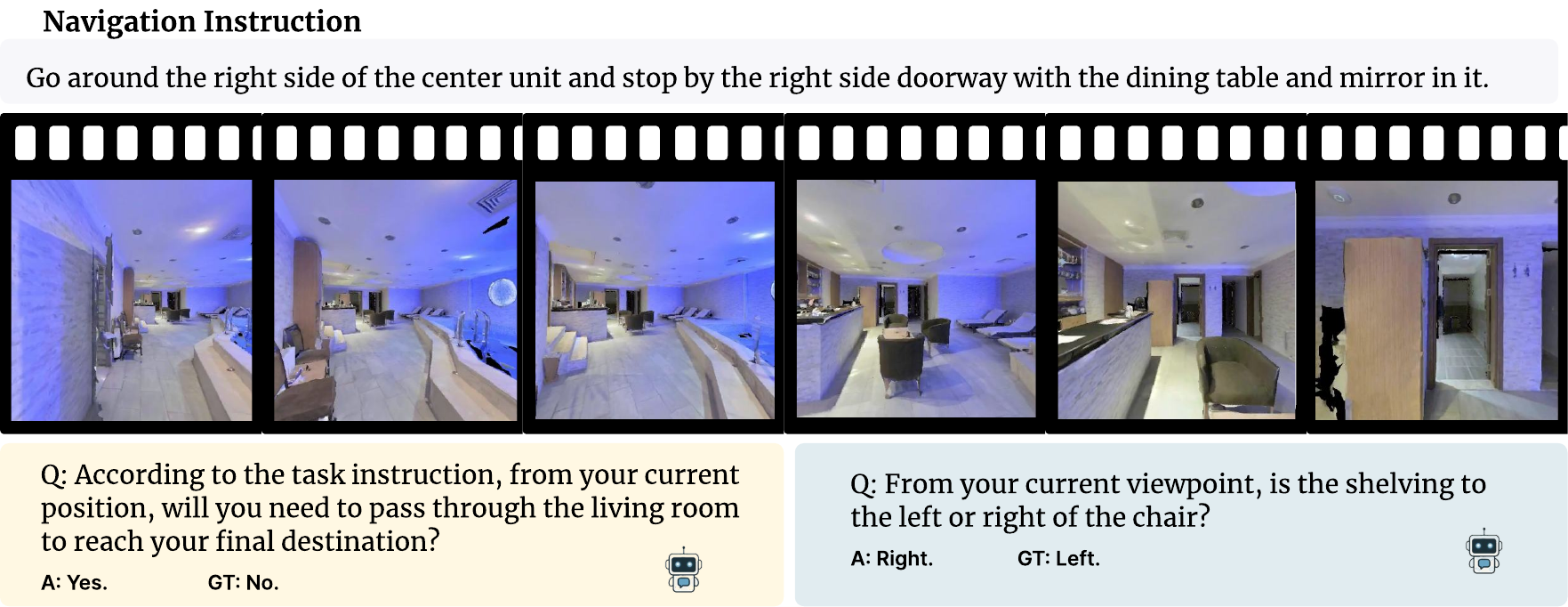}
    \caption{\textbf{Qualitative analysis of reasoning failures.} We illustrate two representative failure modes on the MindCraft-Test set. \textbf{Left (Prospective Failure):} The agent correctly identifies the scene semantics but fails to distinguish the subtle topological difference between ``passing \textit{through}" and ``going \textit{around}" an area. \textbf{Right (Introspective Failure):} The agent struggles with fine-grained geometric grounding, misclassifying the ``Left/Right" relationship due to viewpoint sensitivity.}
    \label{fig:reasoning_failure}
\end{figure}

To provide a more comprehensive understanding of our model's capabilities and limitations, we present additional qualitative examples below.

\paragraph{Success Case.}
The main text has already highlighted our model's ability to follow long and complex instructions, successfully navigating through multiple rooms, while a strong baseline fails by becoming stuck in a local loop.

\subsection{Additional Qualitative Results}
\label{supp:addqual}

To provide a more comprehensive understanding of our model's capabilities and limitations, we present additional qualitative examples below.

\paragraph{Success Case.}
The main text has already highlighted our model's ability to follow long and complex instructions, successfully navigating through multiple rooms, while a strong baseline fails by becoming stuck in a local loop.

\paragraph{Failure Case Visualization and Analysis.}
We further analyze the limitations of our model by categorizing failure modes into navigation execution failures and cognitive reasoning failures.

First, regarding navigation execution, our model occasionally struggles with complex multi-room layouts. As illustrated in Figure~\ref{fig:nav_failure}, we present a failure case with the instruction: ``Walk into the living room and keep walking straight past the living room. Then walk into the entrance under the balcony and wait at the entrance to the other room." Despite the clear visual cues provided in the left image, the model struggles to accurately interpret the global spatial layout, leading to a failure in selecting the optimal path to the destination.

Second, regarding reasoning failures, even when navigation is successful, the model may produce incorrect answers to cognitive queries due to specific cognitive deficits. As visualized in Figure~\ref{fig:reasoning_failure}, we identify two representative reasoning failure modes. The first mode, Topological Ambiguity (Figure~\ref{fig:reasoning_failure} Left), occurs when the model correctly identifies scene semantics but fails to parse fine-grained topological instructions. In the example shown, the instruction commands the agent to ``Go around," but the model incorrectly predicts it must ``pass through" the living room. This suggests a failure to distinguish the subtle linguistic distinction between intersecting a region and skirting its boundary. The second mode, Fine-grained Geometric Grounding (Figure~\ref{fig:reasoning_failure} Right), reveals the model's sensitivity to egocentric viewpoint shifts. When asked about the spatial relationship between the ``shelving" and the ``chair," the model answers ``Right" instead of ``Left." This error highlights the difficulty of resolving precise depth and relative positioning in cluttered scenes where slight camera rotations can invert spatial relationships.




\section{Feature Visualization}
\label{supp:featvis}

To qualitatively validate the structural impact of our ST-CRL objective, we conducted a comprehensive feature space analysis as presented in Figure 4 of the main paper. This visualization comprises two distinct components: a static manifold structure comparison (Top and Middle panels) and a dynamic trajectory evolution analysis (Bottom panel).

For the static manifold analysis, we aimed to demonstrate how ST-CRL shapes the latent space by comparing the learned representations of the baseline model (LASAR IL+QA) against our full model. We randomly sampled $N=1,000$ distinct cognitive map states from the MindCraft-Test validation set for each model. We then performed t-SNE dimensionality reduction independently for each model using a perplexity of 30, a learning rate of 200, and PCA initialization. To interpret the resulting structures, we visualized the same set of projected points under two distinct coloring schemes. The ``Query Type" coloring (Top Row) tests whether the model overfits to linguistic patterns, while the ``Room Type" coloring (Middle Row) evaluates spatial grounding. As observed, the baseline exhibits an isotropic, unstructured distribution under both schemes, indicating a failure to organize memory spatially. In contrast, our model forms dense, distinct clusters when colored by Room Type while maintaining a uniform mix of Query Types within each cluster, confirming that the latent space is organized by spatial semantics rather than superficial query templates.

For the trajectory evolution dynamics shown in the bottom panel, we employed a joint embedding strategy to visualize a continuous navigation episode within the global semantic context. Instead of projecting the trajectory in isolation, we constructed a joint dataset consisting of a background reference set ($S_{\text{ref}}$, $N=2,000$ global validation samples) and a foreground trajectory set ($S_{\text{traj}}$, the sequence of latent states from a single long-horizon episode). We concatenated these sets and performed a single t-SNE run on the combined data to ensure a unified coordinate system. In the final visualization, the background points are rendered with low opacity to delineate the semantic regions (e.g., Kitchen, Bedroom), while the foreground trajectory is connected sequentially using a temporal color gradient. This visualization confirms spatiotemporal consistency, as the agent's projected state smoothly transitions between the corresponding semantic clusters on the manifold as it physically navigates through the environment.

\section{Limitations and Future Work}
\label{sec:limitations_future_work}
While our proposed LASAR framework demonstrates significant advancements in equipping embodied agents with spatio-temporal intelligence, we acknowledge several limitations that pave the way for compelling future research.

First, our model's multi-encoder architecture, while powerful, introduces a notable computational overhead in terms of both memory (VRAM) and floating-point operations (FLOPs) during inference. This currently may hinder its deployment on resource-constrained robotic platforms. Second, the performance of our \textbf{geometric inference module (VGGT)} is contingent on the quality of its \textbf{2D-to-3D pre-training} and may struggle with out-of-distribution visual cues in new environments. The model's robustness against noisy, incomplete, or drifting visual data is an important area for further investigation. Finally, our current framework operates under a static environment assumption. It is not designed to handle dynamic scenes with moving objects, interacting agents, or significant layout changes, which limits its applicability in more complex, human-centric settings.

Addressing these limitations points toward several exciting future directions. A primary focus will be on \textbf{improving model efficiency}. We plan to explore model compression techniques, such as knowledge distillation from our larger model to a more compact student network, and post-training quantization to create a ``LASAR-Lite" version without substantially compromising performance. This would be a critical step towards real-world robotic deployment. 

Another key avenue is to \textbf{enhance the model's robustness and generalization}. To bridge the sim-to-real gap highlighted by our data dependency, we aim to train LASAR on a much wider variety of 3D environments, incorporating simulated sensor noise and diverse visual styles. Furthermore, we will investigate advanced learning paradigms, such as meta-learning or online domain adaptation, to enable the agent to rapidly fine-tune its world model when introduced to a completely new and unseen environment.

Perhaps the most significant long-term vision is to \textbf{transcend the offline training paradigm towards interactive and lifelong learning}. This involves developing mechanisms for the agent to continuously update its semantic and episodic memories based on its own experiences and interactions within the world. By integrating reinforcement learning principles, the agent could learn from trial-and-error, associate failed plans with specific environmental states, and implicitly refine its understanding of physical affordances. Such a system would move us closer to creating truly adaptive and autonomous embodied intelligences that learn and grow over their entire operational lifetime.




\end{document}